\documentclass[11pt]{article}
\usepackage{Frontiers2021}
\colingfinalcopy
\usepackage{times}
\usepackage{url}
\usepackage{latexsym}
\usepackage{booktabs}

\usepackage{graphicx}
\usepackage{tabularx}
\usepackage{comment}
\usepackage{todonotes}
\usepackage{pgfplots}
\pgfplotsset{compat=1.14}
\usepackage{pgfplotstable}

\definecolor{aqua}{rgb}{0.0, 1.0, 1.0}

\newlength\graphheight
\newlength\graphwidth
\pgfplotsset {
graph/.style={
height=\graphheight,
width=\graphwidth,
legend style={
    draw=none,
    fill=none,
    font=\footnotesize,
},
ticklabel style={font=\scriptsize, /pgf/number format/fixed},
scaled y ticks = false,
scaled x ticks = false,
xlabel near ticks,
ylabel near ticks,
},
graphright/.style={
graph,
yticklabel pos=right,
},
}

\title{Neural Machine Translation of Clinical Text: An Empirical Investigation into Multilingual Pre-Trained Language Models and Transfer-Learning}

\author{
          Lifeng Han$ ^1$, Serge Gladkoff$ ^2$,  Gleb Erofeev$^2$ \\ \textbf{Irina Sorokina}$^2$,
          \textbf{Betty Galiano}$^3$, 
\and \textbf{Goran Nenadic}$^{1}$ \\
         $^1$ The University of Manchester, UK \\ 
         $^2$ Logrus Global,  Translation \& Localization \&
         $^3$ Ocean Translations
         \\ {\tt lifeng.han, g.nenadic@manchester.ac.uk} 
         \\
         {\tt 
        serge.gladkoff, gleb.erofeev, irina.sorokina@logrusglobal.com 
        } \\{\tt betty.galiano@oceantranslations.com} \\        }
\date{}

\begin{document}
\maketitle

\begin{abstract}
Clinical texts and documents contain a wealth of information and knowledge in the field of healthcare, and their processing, using state-of-the-art language technology, has become very important for building intelligent systems capable of supporting healthcare and providing greater social good. This processing includes creating language understanding models and translating resources into other natural languages to share domain-specific cross-lingual knowledge. In this work, we conduct investigations on clinical text machine translation by examining multilingual neural network models using deep learning methods such as Transformer-based structures. Furthermore, to address the issue of language resource imbalance, we also carry out experiments using a transfer learning methodology based on massive multilingual pre-trained language models (MMPLMs). The experimental results on three sub-tasks including 1) clinical case (CC), 2) clinical terminology (CT), and 3) ontological concept (OC) show that our models achieved top-level performances in the ClinSpEn-2022 shared task on English-Spanish clinical domain data. Furthermore, our expert-based human evaluations demonstrate that the small-sized pre-trained language model (PLM) wins in the clinical domain fine-tuning over the other two extra-large language models by a large margin. This finding has never been previously reported in the field. Finally, the transfer learning method works well in our experimental setting using the WMT21fb model to accommodate a new Spanish language space that was not seen at the pretraining stage within WMT21fb itself – and this deserves further exploration for clinical knowledge transformation, e.g. investigation into more languages. These research findings can shed some light on domain-specific machine translation development, especially in clinical and healthcare fields. Further research projects can be carried out based on our work to improve healthcare text analytics and knowledge transformations. Our data will be openly available for research purposes at \url{https://github.com/HECTA-UoM/ClinicalNMT} 



\end{abstract}


\section{Introduction}
\label{intro}
In recent years, Healthcare Text Analytics (HECTA) have gained more attention from researchers across different disciplines, due to their impact on clinical treatment, decision-making, hospital operation, and their recently empowered capabilities. These developments have much to do with the latest development of powerful language models (LMs), advanced machine-learning (ML) technologies, and increasingly available digital healthcare data from the social media \cite{griciūtė2023topic,oyebode2021health_social,DL4clinical_social} and discharged outpatient letters from hospital settings \cite{henry20202018_n2c2_task2,spasic2020clinical_review,percha2021modern_clinicalTM}. 

Intelligent healthcare systems have been deployed in some hospitals to support the clinicians' diagnostics and decision-making regarding patients and their health problems \cite{noor2022deployment_TA_hospital,qian2021cpas_UK_ML_hospital}. Such usages include key information extraction (IE) from electronic health records (EHRs), 
normalisation to medical terminologies,
knowledge graph (KG) construction, and relation extraction (RE) between symptoms (problems), diagnoses, treatments, and adverse drug events \cite{wu2022cross_PLM_clinical,nguyen2023spanbased_NER,belkadi2023generating}.
Some of these digital healthcare systems can also help patients self-diagnose in situations where no General Practitioners (GPs) and professional doctors are available \cite{wroge2018parkinson_diagnosis_ML,zhu2021classification_covid_DL}.

However, due to the language barriers and inequal accessibility of digital resources across languages, there is an urgent need for knowledge transfer, such as from one human language to another \cite{costa2022no_NLLB,khoong2022research_agenda_MT4clinical}. 
Thus, to help address digital health disparity, machine translation (MT) technologies can be of good use.

MT is one of the earliest artificial intelligence (AI) branches dating back to the 1950s, and it has boomed in recent years along with other natural language processing (NLP) tasks due to the newly designed powerful Transformers learning model  \cite{Weaver1955,google2017attention,bert2018devlin,han-etal-2021-chinese}. 
Several attention mechanisms designed in Transformer deep neural models have proven themselves capable of better learning from a large amount of available digital data compared to traditional statistical and neural network-based models  \cite{han2022investigation,kuang-etal-2018-attentionNMT,HanKuang2018NMT}.

\begin{figure*}[!t]
\begin{center}
\centering
\includegraphics*[width=0.9\textwidth]{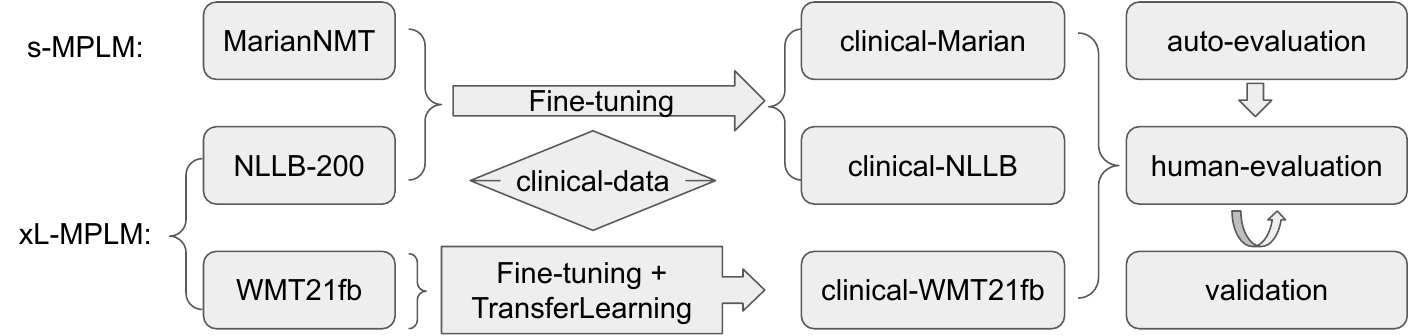}
\caption{Illustration of the Investigation Workflow}
\label{fig:pipeline-clinicalNMT}
\end{center}
\end{figure*}

In this work, we investigate the state-of-the-art Transformer-based Neural MT (NMT) models in connection with clinical domain text translation, to facilitate digital healthcare and knowledge transfer with the workflow drawn in Figure \ref{fig:pipeline-clinicalNMT}.
Being aware of some current development in the competition of language model sizes in the NLP field, we set up the following base models for comparison study: 1) a small-sized multilingual pre-trained Marian language model (s-MPLM), which was developed by researchers  
at the Adam Mickiewicz
University in Poznan and by the NLP group at the University of Edinburgh
\cite{mariannmt,junczys-dowmunt-etal-2018-marian-cost_WNMTG}; and 2) a massive-sized multilingual pre-trained NLLB LM (MMPLM/xL-MPLM), developed by Meta-AI covering more than 200 languages \cite{costa2022no_NLLB}.
In addition to this, we set up a third model to investigate the possibility of transfer learning in the clinical domain MT: 3)  the WMT21fb model which is another MMPLM from Meta-AI but with a limited amount of pre-trained language pairs including from English to Czech, German, Hausa, Icelandic, Japanese, Russian, and Chinese, and the opposite \cite{tran2021facebook}.  

The testing language pairs of these translation models in our work are English $\leftrightarrow$ Spanish. 
As far as we know, there are no other language pairs of openly available resources in the clinical domain MT. 
We use the international shared task challenge data from ClinSpEn2022 ``clinical domain Spanish-English MT 2022'' for this purpose \footnote{\url{https://codalab.lisn.upsaclay.fr/competitions/6696}}. ClinSpEn2022 was a sub-task of the BioMedical MT track at WMT2022 \cite{neves-etal-2022-findings_biomedMT}. 
There are three translation tasks inside ClinSpEn2022 including i) clinical cases report; ii) clinical terms, and iii) ontological concepts from the biomedical domain.

Regarding the evaluation of these LMs, we used the evaluation platform offered by the ClinSpEn2022 shared task including several automatic metrics such as BLEU, METEOR, ROUGE, COMET. However, the automatic evaluation results did not give any apparent differentiation between the models on some of the tasks. Furthermore, there are issues like inconsistency regarding model ranking across automatic metrics. 
To address these issues and give a high-quality evaluation, we performed an expert-based human evaluation on the three models using outputs of Task one ``clinical case report''. 

Our experimental investigation shows that 1) the extra-large MMPLM does not necessarily win over the small-sized MPLM on clinical domain MT via fine-tuning; 2) our transfer-learning model works successfully for clinical domain MT task on language pairs that were not pre-trained for, but added with fine-tuning. 
The first finding can shed some light on the idea that in clinical domain-specific MT, it is better to do more data cleaning and fine-tuning rather than build extra large LMs. 
Our second finding tells us the capability of MMPLMs in generating a new language pair knowledge space for translating clinical domain text even though this language pair was unseen in the pre-training stage with our experimental settings. This can be useful to low-resource NLP, such as the work by \cite{Almansor2018NMT_lowRes,islam2021towards_MT_lowRes}. \footnote{This paper reports systematic investigation findings  based upon the  preliminary work from \cite{han-etal-2022-examining,https://doi.org/10.48550/arxiv.2210.06068zero-shotNMTclinical,han-etal-2023-investigating}.}

The rest of this article is organised as below: Section \ref{relatedwork_section} surveys other works related to ours, including clinical domain MT and NLP, large LMs, and transfer learning. Section \ref{section_model_idea} details the three LMs we deployed for comparison study. Section \ref{model_eval_section} introduces the experimental work we carried out and automatic evaluation outcomes. Section \ref{sec_humanEval} follows up with expert-based human evaluation and the results. Finally, Section \ref{discussion_conclusions} concludes our work with discussion.

\section{Related Work}
\label{relatedwork_section}
Applying NLP models to clinical healthcare has attracted much attention of researchers, such as the work on disease status prediction using discharge summaries by \newcite{yang2009text_clinical},
temporal expressions and events extraction from clinical narratives using combined methods of rules and machine learning by \newcite{kovavcevic2013combining_clinical} and recent deep-learning models by \cite{tu2023extraction}, Temporal Relation modelling on treatments using prompt engineering on GPT models by \cite{cui-etal-2023-medtem2},
using knowledge-based and data-driven methods for de-identification task in clinical narratives by \newcite{dehghan2015combining_de-id}, systematic reviews on clinical text mining and healthcare by \newcite{spasic2020clinical_review} and \newcite{elbattah2021role_health_review}, etc.

However, using MT to help translate clinical text for knowledge transfer and improved clinical decision-making is still a relative novelty \cite{khoong2022research_agenda_MT4clinical}, even though it has  proven its usefulness for assisting \textit{health communication} especially with post-editing strategies \cite{dew2018development_MT_health}. This is partially the result of the sensitive nature of domain and high risk in clinical settings \cite{randhawa2013using_MT_clinical}.  
Some of the recent progress on using MT for clinical texts includes the work by \newcite{soto2019leveraging_SNOMED_ML} which leverages SNOMED-CT terms \cite{donnelly2006snomed} and relations for MT between Basque and Spanish languages; \newcite{mujjiga2019identifying_semantic_MT_clinical} which applies NMT model to identify semantic concepts in 
``abundant interchangeable words'' in clinical domain and their experimental result shows that NMT model can greatly improve the efficiency on extracting
 UMLS \cite{bodenreider2004unified_UMLS} concepts from a single document by using 30 milliseconds in comparison to traditional regular expression-based methods which take 3 seconds; and \newcite{finley-etal-2018-dictations_clinicalMT} which uses NMT to 
 simplify the typical multi-stage workflow on clinical report dictation and even correct the errors from speech recognition. 

With the prevalence of multilingual PLMs (MPLMs) developed from NLP fields, it becomes necessary to test their performances in the clinical domain of NMT.
MPLMs have been adopted for many NLP tasks since the first emergence of the Transformer-based learning structure \cite{google2017attention}. 
Among these, Marian is a small-sized MPLM led by Microsoft Translator based upon Nenatus NMT \cite{nematus} with around 7.6 million parameters \cite{mariannmt}. At the same time, different research and development teams have been competing in recent years in terms of the size of their LMs such as the massive MPLMs (MMPLMs) WMT21fb and NLLB by Meta-AI that have the number of parameters set at 4.7 billion and 54 billion respectively \cite{tran2021facebook,costa2022no_NLLB}.
To investigate the performances of these different models with varied model sizes towards clinical domain NMT with fine-tuning, we set up all three of these as our base models.
To the best of our knowledge, our work is the first to compare small-size and extra-large MPLMs in the clinical domain of NMT.

Close to the clinical domain, there is a biomedical domain MT challengethat has been organised along with the Annual Conference of MT (WMT) since 2016 \cite{bojar-etal-2016-findings_wmt,yeganova-etal-2021-findings_biomedMT}. The historical biomedical MT tasks have covered corpus of biomedical terminologies, scientific abstracts from Medline, summaries of proposals for animal experiments, etc. In 2022, it was the first time that this Biomedical-MT shared task introduced clinical domain data for Spanish-English language pairs \cite{neves-etal-2022-findings_biomedMT}.

As the WMT21fb model does not include Spanish in its pre-training, we also examined the transfer learning technology into the clinical domain NMT towards Spanish-English using the WMT21fb model. 
Transfer-learning \cite{alyafeai2020survey_TransferL} has proved useful for text classification and relation extraction \cite{pomares2021transfer_sp_en_classify_clinical,peng-etal-2019-transfer_biomedNLP}, and low-resource MT \cite{jiang2022transfer_L_LowResMT} fields. 
However, to the best of our knowledge, we are the first to test clinical domain NMT via transfer learning using MMPLMs.

\section{Experimental Designs}
\label{section_model_idea}

In this section, we introduce more information about the three MPLMs that we investigate in this work, i.e., Marian \cite{mariannmt}, WMT21fb \cite{tran2021facebook}, and NLLB \cite{costa2022no_NLLB}.

\begin{figure*}[!ht]
\begin{center}
\centering
\includegraphics*[width=0.80\textwidth]{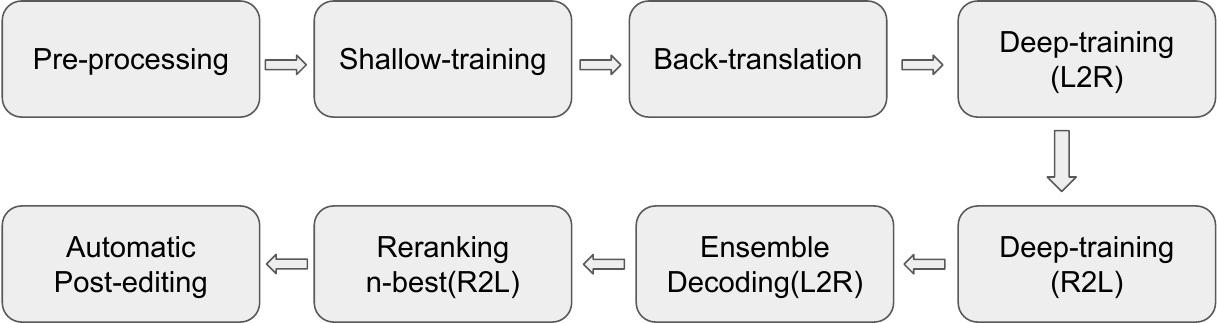}
\caption{Marian Pre-Trained NMT - Training Pipeline. }
\label{fig:Marian_NMT_training_diagram}
\end{center}
\end{figure*}

\begin{figure*}[!ht]
\begin{center}
\centering
\includegraphics*[width=0.6\textwidth]{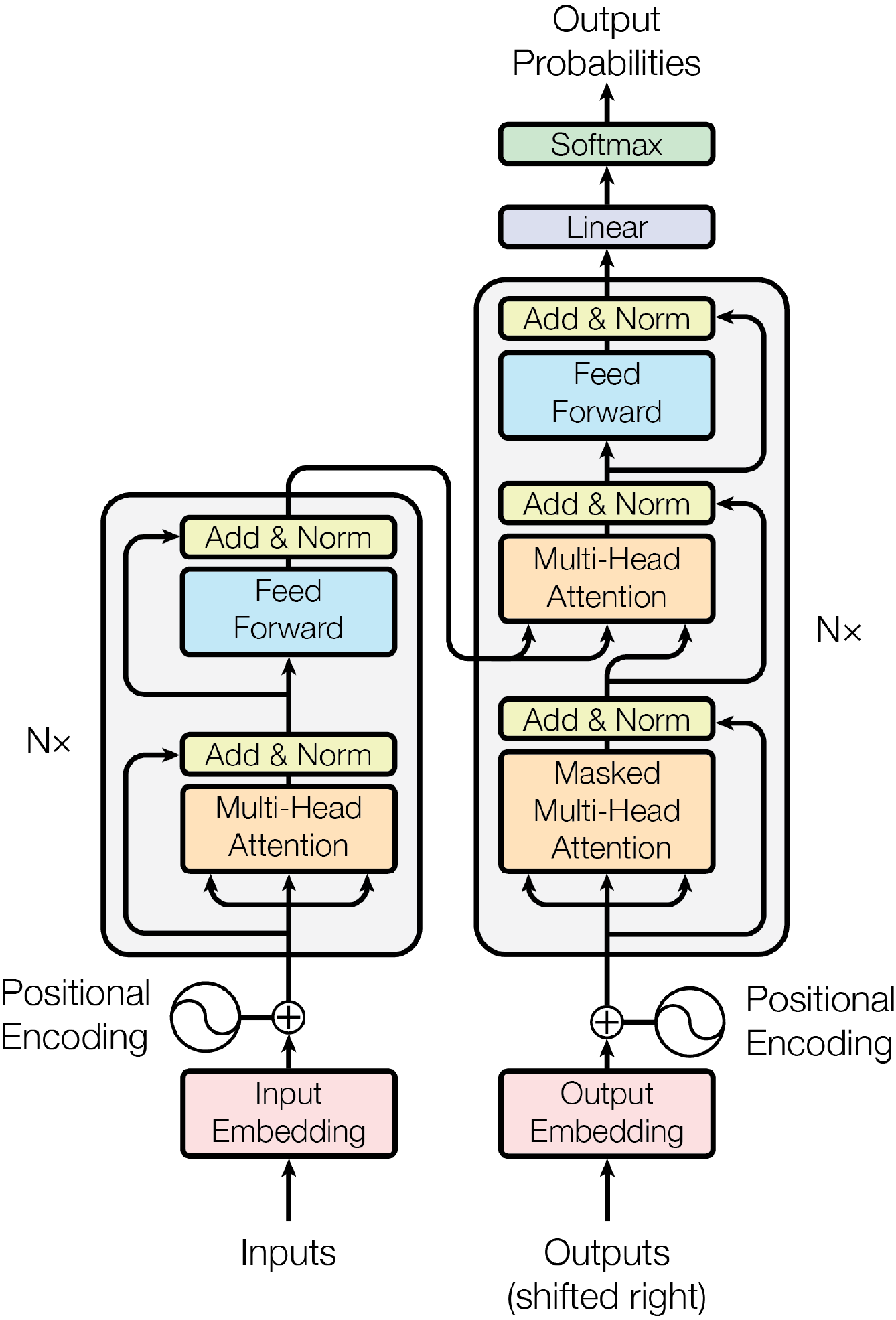}
\caption{Several Attention-based Transformer NMT structure \cite{google2017attention}. }
\label{fig:transformer_fig}
\end{center}
\end{figure*}

\begin{figure*}[!t]
\begin{center}
\centering
\includegraphics*[width=0.8\textwidth]{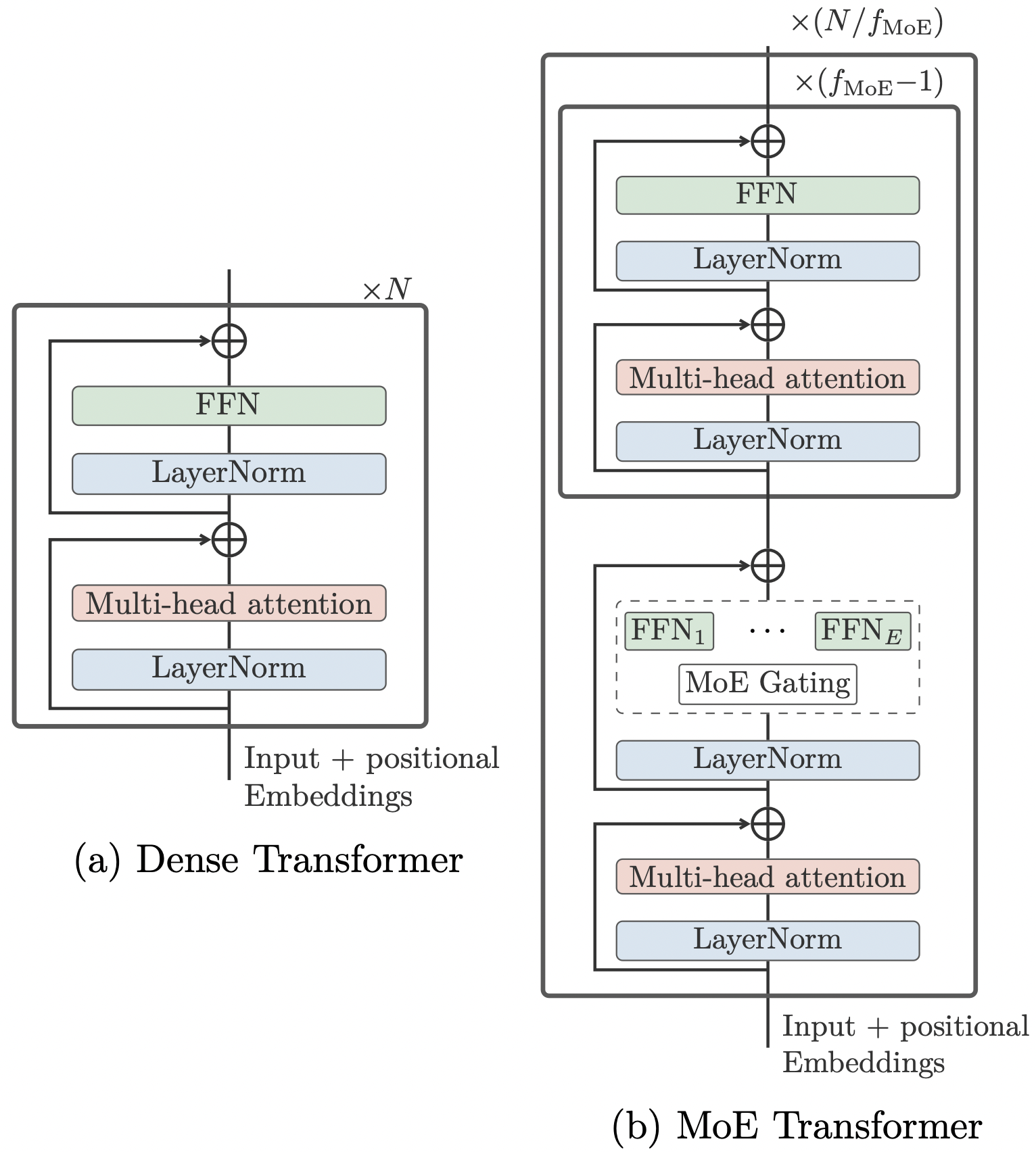}
\caption{Original Dense Transformer vs MoE Transformer \cite{costa2022no_NLLB}}
\label{fig:dense_vs_MoE_Transformer}
\end{center}
\end{figure*}

\begin{figure*}[!t]
\begin{center}
\centering
\includegraphics*[width=0.8\textwidth]{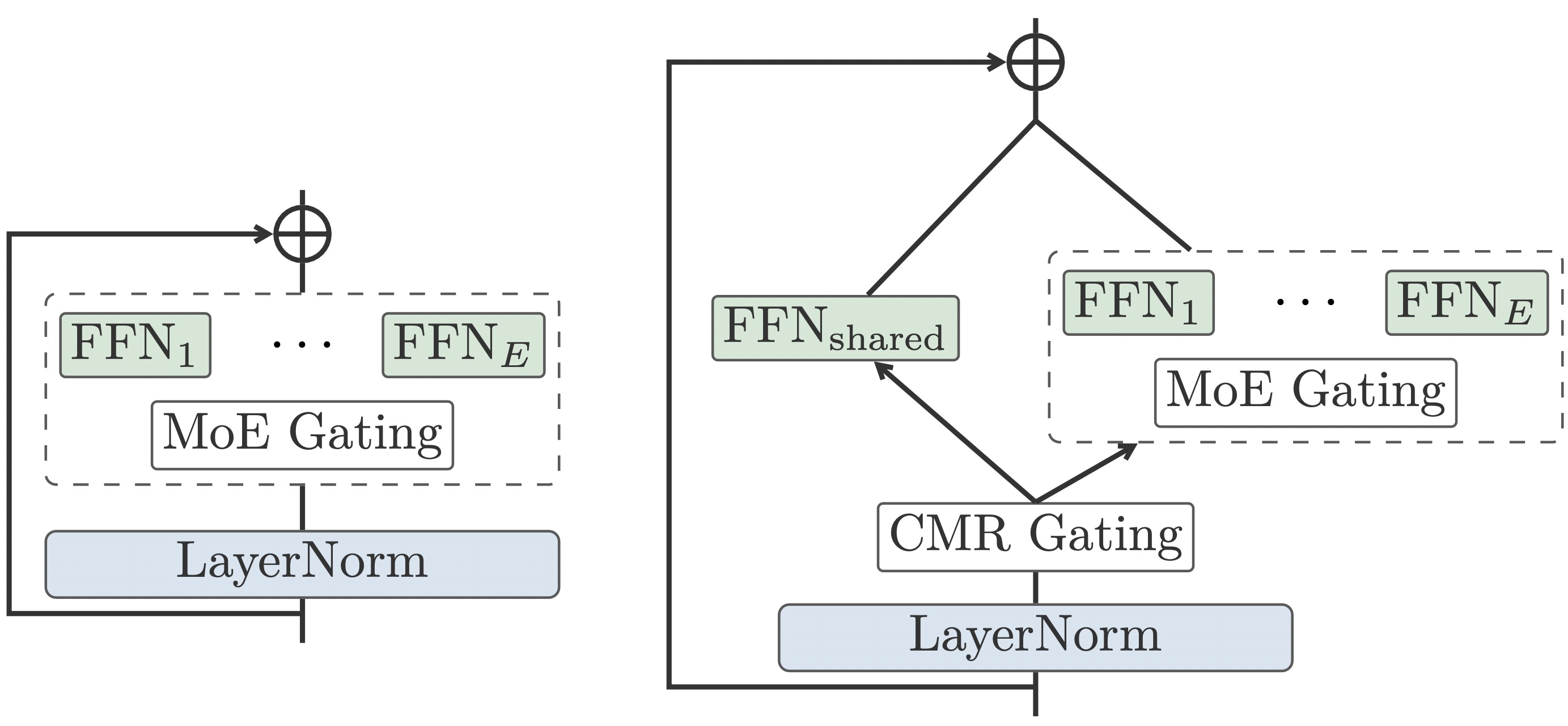}
\caption{MoE vs Conditional MoE \cite{costa2022no_NLLB}}
\label{fig:MoE_vs_ConditionalMoE}
\end{center}
\end{figure*}

\subsection{Multilingual Marian NMT}
First, we draw a training diagram of the original Marian model on its pre-training steps in Figure \ref{fig:Marian_NMT_training_diagram} according to \cite{mariannmt}. The pre-processing step includes tokenisation, true-casing, and Byte-Pair Encoding (BPE) for sub-words. The shallow training is to teach a mid-phase translation model to produce temporary target outputs for back-translation. Then, the back-translation step produces the same amount of input source sentences to enlarge the corpus. 
The deep-training step first uses four left-to-right models which can be RNN \cite{nematus} or Transformer \cite{google2017attention} structures, which is followed by four right-to-left models in the opposite direction. 
The ensemble-decoding step will generate the n-best hypothesis translations for each source input segment, which will be re-ranked using a re-scoring mechanism. 
Finally, in Marian NMT, there is an automatic post-editing step taken before the final output is produced. This step is also based on an end-to-end neural structure by modelling the \textit{set}(MT-output, source sentence)$\rightarrow$``post-edited output'' as introduced by \newcite{junczys-dowmunt-grundkiewicz-2017-exploration_auto-PED}.

The Marian NMT model we deployed is from the Language Technology Research Group at the University of Helsinki led by \newcite{TiedemannThottingal:EAMT2020_OPUS_MT}. It is based on the original Marian model but continuously trained on the multilingual OPUS corpus \cite{tiedemann2012parallel_OPUS} to make the model available to more languages. It includes Spanish$\leftrightarrow$English (es$\leftrightarrow$en) pre-trained models and has
7.6 million parameters for fine-tuning. \footnote{\url{https://huggingface.co/Helsinki-NLP}} 

\subsection{Extra-Large Multilingual WMT21fb and NLLB}

Instead of the optional RNN structure used in the Marian model, both WMT21fb and NLLB massive-sized multilingual PLMs (MMPLMs)  adopted Transformer as the main methodology. As shown in Figure \ref{fig:transformer_fig}, Transformer's main components for encoder include position encoding, Multi-Head Attention, and Feed-Forward Network with layer normalisation at both two steps. The decoder uses Masked Multi-Head Attention to constrain the generation model only taking into account the already generated text. 

To increase the model capacity without making the extra-large model too slow for training, the WMT21fb model included ``Sparsely Gated Mixture-of- Expert (MoE)'' models, 
inspired by the work from  \newcite{lepikhingshard_Scaling}, into the FFN layer of the Transformer, as shown in Figure \ref{fig:dense_vs_MoE_Transformer}. The MoE model will only pass a sub-set of model parameters into the next level, thus decreasing the computational cost. However, this dropout is done in a random manner.

Furthermore, this structure design still needs language-specific training, such as English-to-other and other-to-English used by WMT21fb.

To further improve on this, the NLLB model designed a Conditional MoE Routing layer inspired by \newcite{zhang2021share_schedule} to ask the MoE model to decide which tokens to dropout based on how computationally intensive/resource-heavy they are to process or based on their routing efficiency.
This is achieved by a binary gate, which assigns weights to dense FNN $FFN_{shared}$ or MoE Gating, as in Figure \ref{fig:MoE_vs_ConditionalMoE}.
The Conditional MoE also 
removes language-specific parameters for learning.

In summary, the WMT21fb and NLLB models share very similar learning structures, the biggest difference being that WMT21fb used language-specific constrained learning.
The WMT21fb model we applied is `wmt21-dense-24-wide.En-X' (and X-En direction) which has 4.7 billion parameters \footnote{\url{https://github.com/facebookresearch/fairseq/tree/main/examples/wmt21}} and contains the language pairs English $\leftrightarrow$ Chinese, Czech, German, Hausa, Icelandic, Japanese, and Russian.
The full NLLB model includes 200+ languages and has 54.5 billion parameters. 
Due to the computational restriction, we applied the distilled model of NLLB, i.e. NLLB-distilled, which has 1.3 billion parameters.

The WMT21fb model does not have Spanish among its trained language pairs, while NLLB includes Spanish as a high-resource language. This is a perfect setting for us to examine the transfer-learning technology on the clinical domain NMT by fine-tuning a translation model for the Spanish language on the WMT21fb model and comparing the output with the NLLB model (Spanish version).

\section{Experimental Settings and Evaluations}
\label{model_eval_section}


\begin{table*}[!t]
\begin{center}
\centering
\begin{tabular}{ccccccc}
\toprule
\multicolumn{1}{c}{} 
     & \multicolumn{5}{c}{Task-I: Clinical Cases (CC) EN$\rightarrow$ES}    \\ \hline 
\multicolumn{1}{c}{MT fine-tuning} 
     & plm.es & \multicolumn{1}{c}{S\textsc{acre}BLEU}     
                & METEOR & COMET & BLEU & ROUGE-L-F1 \\
\midrule
Clinical-Marian & Yes & \textit{38.18} &\textit{0.6338} &\textit{0.4237} &\textit{0.3650} &\textit{0.6271}  \\
Clnical-NLLB & Yes & 37.74 & 0.6273& 0.4081& 0.3601& 0.6193 \\
Clinical-WMT21fb & \textbf{No} &34.30 &0.5868 &0.3448 &0.3266 & 0.5927\\
\hline\hline
\multicolumn{1}{c}{} 
     & \multicolumn{5}{c}{Task-II: Clinical Terms (CT)  EN$\leftarrow$ES }    \\ \hline 
\multicolumn{1}{c}{MT fine-tuning} 
     & plm.es & \multicolumn{1}{c}{S\textsc{acre}BLEU}     
                & METEOR & COMET & BLEU & ROUGE-L-F1 \\
\midrule
Clinical-Marian & Yes & 26.87 &\textit{0.5885} &0.9791 &0.2667 &\textit{0.6720} \\
Clinical-NLLB & Yes& \textit{28.57} & 0.5873 & \textit{{1.0290}} & \textit{{0.2844}} & 0.6710 \\
Clinical-WMT21fb & \textbf{No} &24.39 &0.5840 &0.8584 & 0.2431 & 0.6699\\
 \hline\hline
\multicolumn{1}{c}{} 
     & \multicolumn{5}{c}{Task-III: Ontology Concept (OC)  EN$\rightarrow$ES  }    \\ \hline
\multicolumn{1}{c}{MT fine-tuning} 
     & plm.es & \multicolumn{1}{c}{S\textsc{acre}BLEU}     
                & METEOR & COMET & BLEU & ROUGE-L-F1 \\
\midrule
Clinical-Marian & Yes &39.10 & \textit{0.6262} & {0.9495} & 0.3675 &\textit{0.7688} \\
Clinical-NLLB & Yes& \textit{{41.63}}  & 0.6072& 0.9180& \textit{{0.3932}} & 0.7477 \\
Clinical-WMT21fb& \textbf{No} &40.71 &0.5686  & \textit{0.9908} & 0.3859 & 0.7199 \\
\bottomrule
\end{tabular}
\caption{Automatic Evaluation of Three MPLMs using ClinSpEn-2022 Platform. `plm.es' means if the Spanish language is included in PLMs.}
\label{tab:clinSpEn_eval_score_t123}
\end{center}
\end{table*}

\subsection{Domain Fine-tuning Corpus}
To fine-tune the three MPLMs for English $\leftrightarrow$ Spanish language pair towards the clinical domain, we used the medical bilingual corpus MeSpEn from \newcite{villegas2018mespen}, which contains sentences, glossaries, and terminologies. 
We performed data cleaning and extracted around 250K pairs of segments in this language pair for domain fine-tuning of the three models. These extracted 250K pairs of segments are randomly chosen from the original MeSpEn corpus and we divided them into 9:1 ratio for training and development purposes. 
Because the WMT21fb pre-trained model did not include Spanish as one of the pre-trained language models, we could not use $<2es>$ (to-Spanish) indicator for fine-tuning. As a solution, we used $<2ru>$ as the indicator for this purpose (to-Spanish).
This means a transfer learning challenge to investigate if the extra-large multilingual PLM (xL-PLM) WMT21fb has created a semantic space to accommodate a new language pair for translation modelling using the 250K size of corpus we extracted.

\subsection{Model Parameter Settings}
\begin{figure*}[!t]
\begin{center}
\centering
\includegraphics*[width=0.8\textwidth]{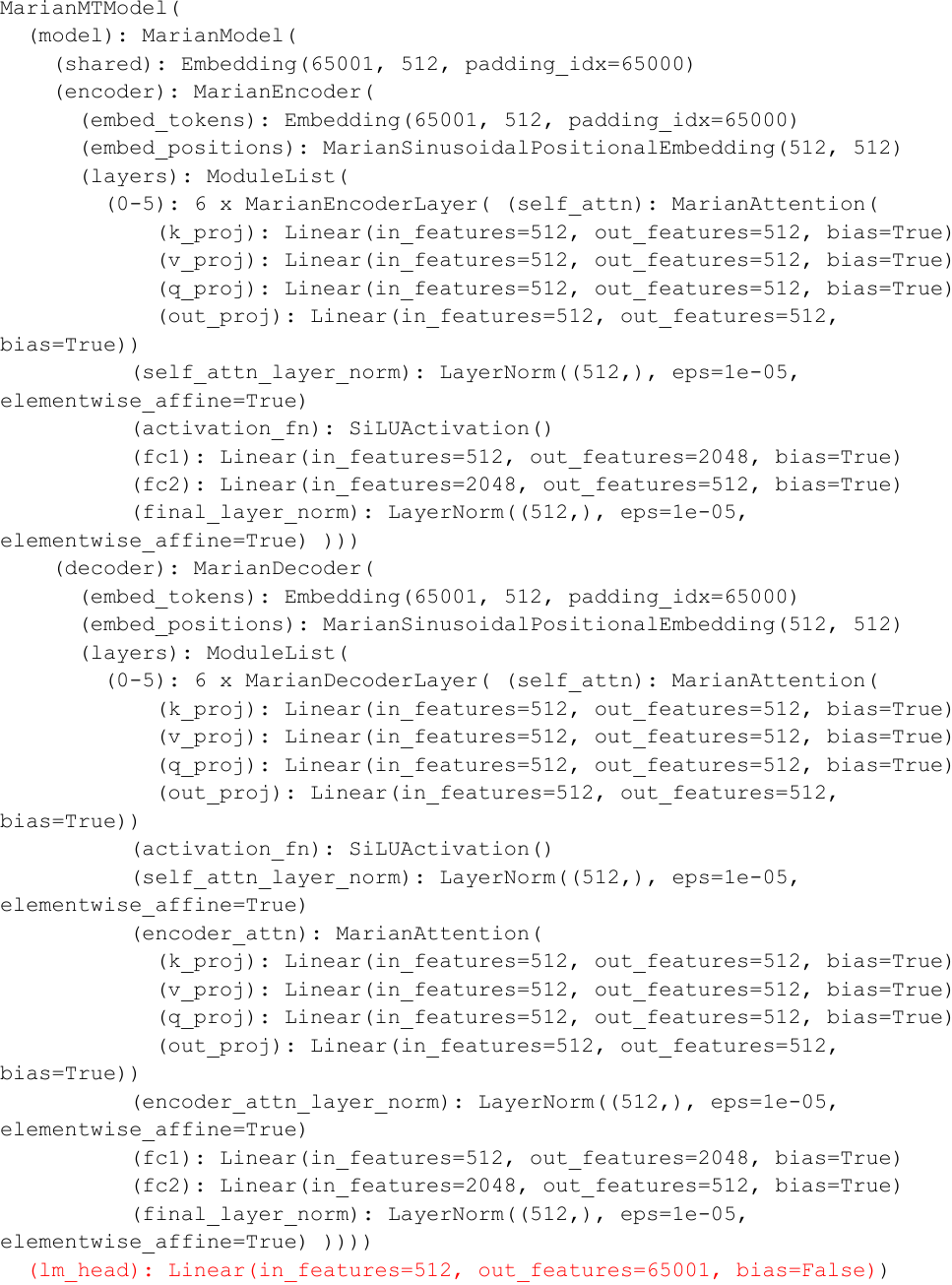}
\caption{MarianNMT Fine-Tuning Parameters: Encoder and Decoder with 6+6 Layers}
\label{fig:MarianMTModel_param_stru}
\end{center}
\end{figure*}

Some parameter settings for s-MPLM Marian model fine-tuning are listed below. The last activation function for generative model is a linear layer. Within the decoder and encoder, we used the Sigmoid Linear Units (SiLU) activation function. More detailed parameter and layer settings are displayed in Figure \ref{fig:MarianMTModel_param_stru} (Appendix).

\begin{itemize}
    \item learning rate = 2e-5
    \item batch size = 128
    \item weight decay - 0.01
    \item training epochs = 1
    \item encoder-decoder layers = 6+6
\end{itemize}

\noindent Some fine-tuning parameters for NLLB-200-distilled \cite{costa2022no_NLLB} are listed below:
\begin{itemize}
    \item batch size = 24
    \item gradient accumulation steps = 8
    \item weight decay = 0.01
    \item learning rate = 2e-5
    \item Activation function (encoder/decoder) = ReLU
    \item number of training epochs = 1
    \item encoder-decoder layers = 24+24
\end{itemize}

\begin{figure*}[!t]
\centering
\includegraphics*[width=0.85\textwidth]{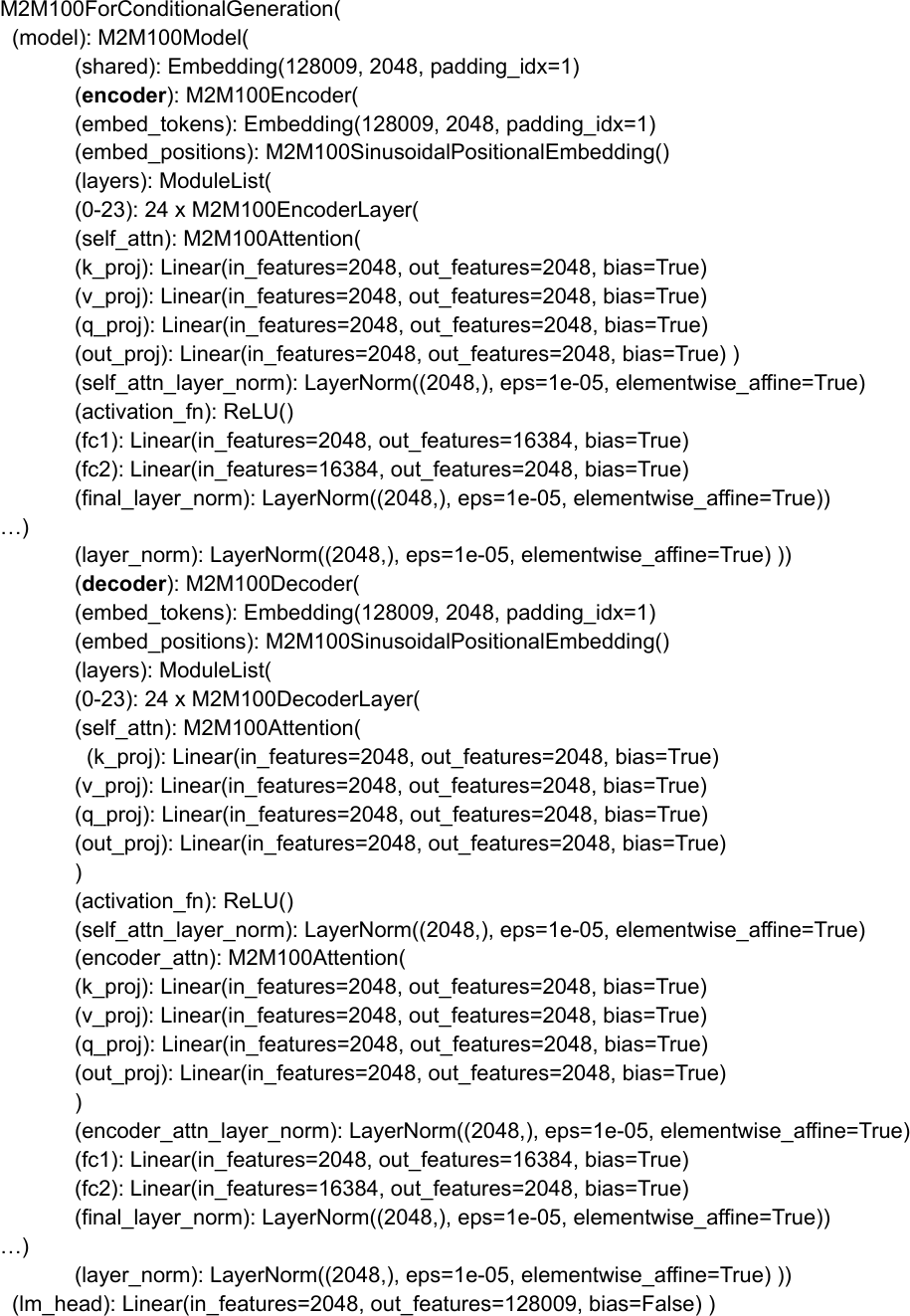}
\caption{M2M-100 Model Structure For Conditional Generation: Encoder and Decoder Parameters with 24+24 Layers}
\label{fig:M2M_100_encoder}
\end{figure*}

\noindent The fine-tuning parameters for WMT21fb model are the same as the NLLB-200-distilled, except for the batch size value which is set as 2. This is because the model is too large and we would get out-of-memory (OOM) errors if we increase the batch size to anything larger than 2.
\noindent More details on M2M-100 parameters and layer settings for Conditional Generation Structure \cite{10.5555/3546258.3546365_m2m_100} we used for xL-MPLM WMT21fb and NLLB-200
can be found in Figure \ref{fig:M2M_100_encoder} (Appendix).

\subsection{Test Sets and Automatic Evaluations}
The evaluation corpus we used is from the ClinSpEn-2022 shared task challenge data organised as part of the Biomedical MT track in WMT2022 \cite{neves-etal-2022-findings_biomedMT}.
It has three sub-tasks: 1) EN$\rightarrow$ES translation of 202 COVID19 clinical case reports; 
2) ES$\rightarrow$EN translation of 19K clinical terms  from biomedical literature and EHRs; and 3) EN$\rightarrow$ES 2K ontological concept from biomedical ontology.

The automatic evaluation metrics used for testing include 
BLEU (HuggingFace) \cite{papineni-etal-2002-bleu}, ROUGE-L-F1 \cite{lin-2004-rouge},
METEOR \cite{BanerjeeLavie2005}, S\textsc{acre}BLEU \cite{post-2018-call4clarity}, and COMET \cite{rei-etal-2020-comet}, hosted by the ClinSpEn-2022 platform \footnote{\url{https://temu.bsc.es/clinspen/}}.
The metric scores are reported in Table \ref{tab:clinSpEn_eval_score_t123} for three translation tasks. In the table, the parameter `plm.es' is a question mark asking if the Spanish language was already included in the original off-the-shelf PLMs. For this question, both Marian and NLLB have Spanish in their PLMs, while WMT21fb does not, which indicates that Clinical-WMT21fb  is a transfer learning model for EN$\leftrightarrow$ES language pair.

From this automatic evaluation result, the first surprising finding is that the much smaller Clinical-Marian model had most of the highest scores across the three tasks, as indicated by \textit{italics}. 
The second finding concerns the two xL-MPLMs: even though the transfer-learning model Clinical-WMT21fb has a certain score gap to Clinical-NLLB on Task 1, it almost catches up with Clinical-NLLB for Task 2 and 3 even winning one of the scores, the COMET for Task 3 (0.9908 vs 0.9180).
This means that the xL-MPLM has the capacity to create a multilingual semantic space and the capability to generate a new language model as long as there is a sufficeint amount of fine-tuning corpus for this new language.
Third, there are issues with automatic metrics. This includes the confidence level on score difference (significance test), such as the very closely related scores for Task 1 on the first two winner models. In addition, the winner models change across Task 2 and 3 via different metrics.

We also observed that there are 4 percent of Russian tokens in the EN $\rightarrow$ ES output from the Clinical WMT21fb model. This indicates that the model keeps Russian tokens when it does not know how to translate the English token into Spanish. This is very interesting since the Russian tokens reserved in the text are not a nonsense - instead, they are tokens with correct meaning, only in a different language. 
This might be the reason why COMET generated higher score for Clinical-WMT21fb model than Clinical-NLLB on Task-3 `ontological concept' since COMET is a neural metric that calculates the semantic similarity on an embedding space, ignoring the word surface form.

To improve the trustworthiness of our empirical investigation and generate a clearer evaluation output across the three models, we perform human expert-based evaluations in the next section.

\subsection{Comparisons}

To compare our much smaller Clinical-Marian model with other existing work on this shared task data, such as Optum \cite{manchanda-bhagwat-2022-optums} and Huawei \cite{wang-etal-2022-huawei}, we list the automatic evaluation scores in Table \ref{tab:clinSpEn_eval_score_t123_offcial_vs_teams} where Optum attended all three sub-tasks, while Huawei only attended Task 2: Clinical Terminology (CT).
From the comparison scores using automatic metrics, we can see that the much smaller Clinical-Marian wins some metrics in each of the tasks. In addition, Optum used their in-house clinical data as extra training resources in addition to WMT-offered training set, while the 250K training set we used for Clinical-Marian is extracted only using WMT data.
Huawei's model only wins one metric (COMET) out of five metrics on Task 2 (CT), however, both Clinical-Marian and Optum win two metrics out of five. 
This means that Huawei's performance  on this task is not much better even though they have much greater online resources and computational support.

\begin{table*}[!t]
\begin{center}
\centering
\begin{tabular}{crcccc}
\toprule
\multicolumn{1}{c}{} 
     & \multicolumn{5}{c}{Task-1: Translating Clinical Cases}    \\ \hline 
\multicolumn{1}{c}{Models} 
     & \multicolumn{1}{c}{S\textsc{acre}BLEU}     
                & METEOR & COMET & BLEU & ROUGE \\
\midrule
Clinical-Marian & {\underline{38.17}} & {0.6337} & {0.4237} & \underline{0.3650} & {0.6270}  \\
Optum & 38.12 & \underline{0.6447} & \underline{0.4425} &0.3642 & \underline{0.6285} \\
\hline
\multicolumn{1}{c}{} 
     & \multicolumn{5}{c}{Task-2: Clinical Terminologies}    \\ \hline 
\multicolumn{1}{c}{Models} 
     & \multicolumn{1}{c}{S\textsc{acre}BLEU}     
                & METEOR & COMET & BLEU & ROUGE \\
\midrule
Optum & \underline{44.97} & 0.5880 & 1.1197 & \underline{0.4396} & 0.7479 \\
Huawei & 41.57 & 0.624 & \underline{1.190}   & 0.4132 &  0.721\\
Clinical-Marian & {{39.10}} & \underline{0.6261} & {0.9494} & {0.3674} & \underline{0.7688}  \\
\hline
\multicolumn{1}{c}{} 
     & \multicolumn{5}{c}{Task-3: Translating Ontology Concepts}    \\ \hline 
\multicolumn{1}{c}{Models} 
     & \multicolumn{1}{c}{S\textsc{acre}BLEU}     
                & METEOR & COMET & BLEU & ROUGE \\
\midrule
Optum & \underline{44.97} & 0.5880 & \underline{1.1197} & \underline{0.4396} & 0.7479 \\
Clinical-Marian & {{39.10}} & \underline{0.6261} & {0.9494} & {0.3674} & \underline{0.7688}  \\
\hline
\bottomrule
\end{tabular}
\caption{Model Comparisons on 3 Tasks between Clinical-Marian and Others.}
\label{tab:clinSpEn_eval_score_t123_offcial_vs_teams}
\end{center}
\end{table*}

\section{Human Evaluation}
\label{sec_humanEval}

As observed in the last section, we had two reasons to set up the expert-based human evaluation: 1) it is really surprising that the much smaller MPLM (s-MPLM) Clinical-Marian performs better than the xL-MPLMs Clinical-NLLB and Clinical-WMT21fb; 2) to verify the automatic evaluation hypothesis that Clinical-Marian really does have the best performance.

\subsection{Human Evaluation Setup}

To achieve both the qualitative and quantitative human evaluation, we deployed a human-centric expert-based post-editing quality evaluation metric called HOPE by \newcite{gladkoff-han-2022-hope} (it is also called LOGIPEM and invented by Logrus Global LLC, a language service provider). 
The HOPE evaluation metric has 8 predefined error types and each error type has corresponding different levels of penalty points according to the severity level. The sentence level and system level HOPE score is a comprehensive score reflecting the overall quality of outputs.

First, we recruited five human evaluators who have the backgrounds in professional translation, linguistics, and biomedical research.
For the evaluation data set, we took all the test set output from Task 1 `clinical case' reports since this is the only task with full sentences. 
For the other two tasks on term and ontology level translation, MT engines can produce relatively good outcomes even without an effective encoder-decoder neural model, e.g. via a well-prepared bilingual dictionary. 
We prepared 100 strings for each set and delivered all the sets to five professional evaluators \footnote{The 100 examples for evaluators were randomly selected from the test dataset and we we will make the data available.}.
The tasks consisted of strings of medical cases going in order one by one, so the context of each case is clear to the evaluator.

Each one of them was given three files for evaluation from different engines, and instructions were given on both the online Perfectionist tool that was used for evaluation and the HOPE metrics. 
Then, to ensure the human evaluation quality, we have also asked the strictest reviewer/evaluator to validate the work of other evaluators.
The strictest reviewer is one of our experts from the language service provider industry and has our trust according to their long-term experiences in post-editing MT outputs and selecting MT engines in real world projects.
The strictest reviewer made better distinctions between all three evaluated models, while the less-strict reviewers sometimes gave similar scores to these models without picking their errors rigorously.

\subsection{Human Evaluation Output}
\begin{figure*}[!t]
\begin{center}
\centering
\includegraphics*[width=0.90\textwidth]{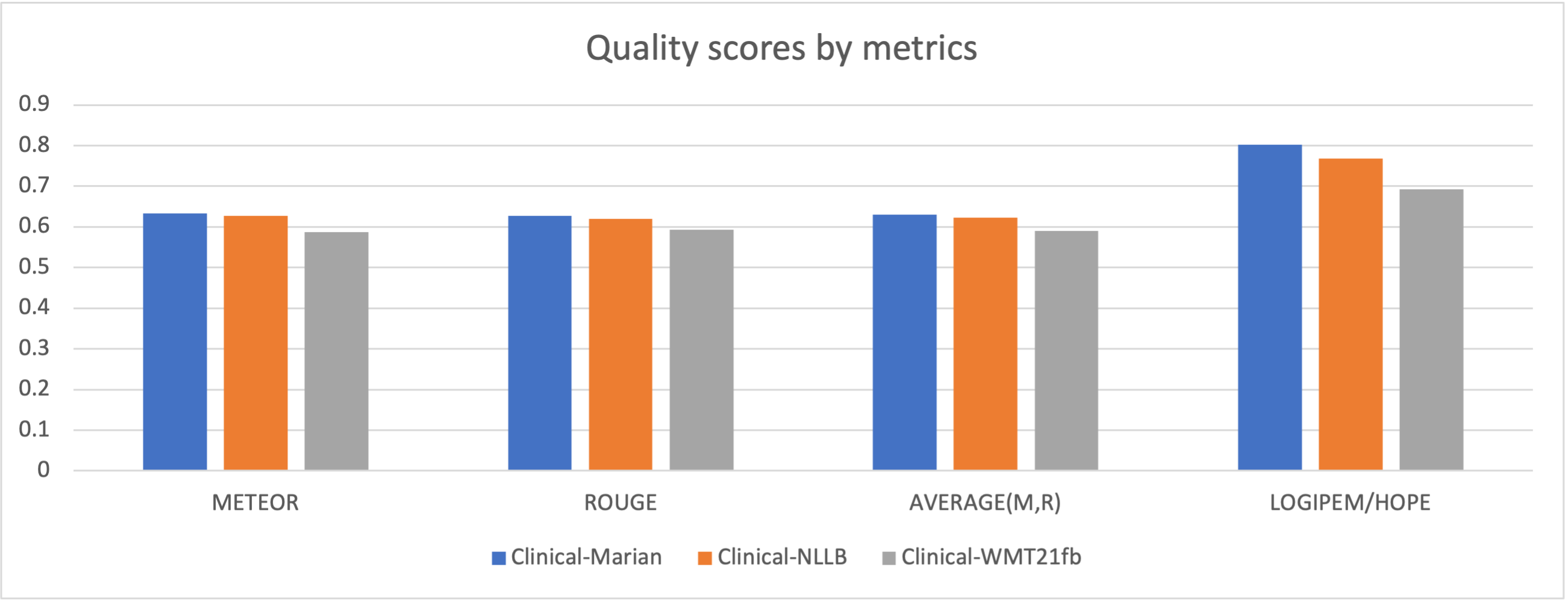}
\caption{Comparison of Automatic Evaluations against Human Evaluation (HOPE)}
\label{fig:compare_auto_human_eval-cropped}
\end{center}
\end{figure*}

The results of the evaluation can be seen in the online Perfectionist tool that was used for this purpose, as downloaded from the tool in the form of familiar Excel scorecards. They are tallied in Figure \ref{fig:compare_auto_human_eval-cropped} and Table \ref{tab:autoEval_vs_HumanEval}.
The human evaluation clearly shows which model is the best demonstrating a large score gap in-between: the Clinical-Marian has a score of 0.801625, followed by Clinical-NLLB and Clinical-WMT21fb with scores of 0.768125 and 0.692429 respectively.

To compare the human evaluation outputs with the automatic metric scores, we also added two metrics, METEOR and ROUGE, and their average score into the figure. 
The reason we chose these two particular metrics is that they have a relatively positive correlation to human judgements.
For the other three metrics, there are several issues that prevented their use. First, BLEU shows NLLB as being better for terms and concepts, which does not correspond to the human judgement. 
Moreover, BLEU shows WMT21fb concepts to be better than those of the Marian Helsinki model, which is completely incorrect.
Second, COMET score for the NLLB model is higher than 1, which is clearly caused by the fact that this implementation of COMET was not normalised by the Sigmoid function. Also, this COMET score for NLLB is higher than the one for Marian Helsinki. Another error is that the COMET score for clinical cases is much better than for both Marian and NLLB, which is completely impossible due to the presence of foreign language tokens in WMT21fb output. 
Finally, when we see COMET scores like 0.99 and 0.949 for Concepts, the score 0.42, 0.40 and 0.34 for Cases look clearly out of line.
The BLEU-HF scores for all content types are ridiculously low on the scale of [0, 1] for both Cases and especially for Terms.

Below is the list of findings made from the comparisons.

\begin{itemize}
    \item Most importantly, all human evaluators consistently showed positive correlation with preliminary human judgement of the MT output quality. Some of them gave more rigorous evaluations than the others, but all of them rated the worst model as the worst and the best model as the best with only one exception. Results of human evaluation fully confirm initial hypothesis about the quality of outputs of different engines, which is based on initial holistic spot-check human evaluation.
    \item The LOGIPEM/HOPE metric shows a much greater difference in output quality than any of the automated metrics. Where the automatic score shows a 6 percent difference, human evaluation gives 14 percent. In other words, the human linguists clearly see a significant difference between output quality of different engines. Even the less-trained evaluators show a positive correlation with the hypothesis.
    \item Even for those automatic metrics that correlate with human judgement, the score values do not seem to be representations of the uniform interval of [0, 1]. The LOGIPEM/HOPE score will be exactly 1 if the segments, in the reviewer’s opinion, do not have to be edited, and LOGIPEM/HOPE score of 0.8 means only about 20\% by total wordcount of work left to be done on the text with that score, since the LOGIPEM/HOPE scoring model is designed with productivity assumptions in mind for various degrees of quality. The COMET or ROUGE score of 0.6 means that MT has generated words that are different from those in the reference, and this in turn means that even a perfect translation which is different from the reference would be rated much lower than 1. This is a huge distortion of linearity, which is metric-specific because all scores for different metrics live in their own ranges. Automatic scores appear to live on some sort of non-uniform scale of their own, which is yet another reason why they are not comparable to each other. The scale is compressed, and the difference between samples becomes statistically insignificant.
    \item The margin of error for all three engines is about 6\%, which is about the same as the difference between the mean of the measurements for different engines. This means that the difference between measurement is statistically significant, but a lot depends on the subjectivity of the reviewer, and the difference between reviewers’ positions may negate the difference in scores. However, even despite the reviewers' subjectivity, the groups of measurements for different engines appear to provide a statistically and visually significant difference.
    \item In general, human evaluators have to be trained / highly experienced, and need to maintain a certain level of rigour. The desired target quality should be stipulated quite clearly by customer specifications, as defined in ISO 11669 and ASTM F2575. To avoid incorrect (inflated) scores and decrease Inter-Rater Reliability (IRR), the linguists must be either tested prior to doing evaluations or cross-validated afterwards.
    \item One evaluation task only takes 1 hour. There were 24 evaluation tasks in total, each task with 100 segments. It does not require setting up any data processing, software development, reference “golden standard” data or model-trained evaluation metric. It is clearly faster, more cost-effective and reliable than the research on whether an automatic metric can even pass the positive correlation test with human judgement (3 out of 5 did not in our case). While individual human measurements have variance, they are all valid and all correlate with human judgement if done with minimal training and rigour.
    \item Automatic metrics are not comparable across different engines, different data sets, different languages and different domains. On the contrary, human measurement is the golden universal standard that provides the least common denominator between these scenarios. In other words, if Rouge is 0.67 for En-Fr for medical text, and Rouge is 0.82 for En-De for automotive text, we can’t compare these numbers. In contrast, LOGIPEM/HOPE score would mean one and the same thing across the board.
\end{itemize}

All of the above confirms the validity and interoperability of our human evaluation using LOGIPEM/HOPE metrics \cite{gladkoff-han-2022-hope}, which can be used as a single quick and easy validator of automatic metrics, and the ultimate fast and easy way to carry out analytic quality measurement to compare the engines and evaluate the quality of translation and post-editing.

\begin{table*}[!t]
\begin{center}
\centering
\begin{tabular}{ccccccc}
\toprule
\multicolumn{1}{c}{MPLMs} 
     & \multicolumn{2}{c}{Auto. Metrics}     
                & \multicolumn{2}{c}{Average}  & \multicolumn{2}{c}{Diff. in Scores} \\\hline
                \multicolumn{1}{c}{} 
     & METEOR & ROUGE     
                & Averge(M,R) & HOPE & Auto. & HOPE \\
\midrule
Clinical-Marian & 0.6338 & 0.6271 & 0.6304 & 0.8016
 &6.45\% &13.62\%  \\
Clnical-NLLB & 0.6273	& 0.6193 &	0.6233	&0.7681	&	1.13\%	& 4.18\%
\\
Clinical-WMT21fb & 0.5868	&0.5927	& 0.5898	&0.6924	&	5.38\%	&9.85\%
\\\hline
\bottomrule
\end{tabular}
\caption{Automatic Evaluations vs Human Evaluations (HOPE) on Three MPLMs}
\label{tab:autoEval_vs_HumanEval}
\end{center}
\end{table*}

\subsection{Inter-Rater-Reliability}

\begin{figure*}[!t]
\begin{center}
\centering
\includegraphics*[width=0.90\textwidth]{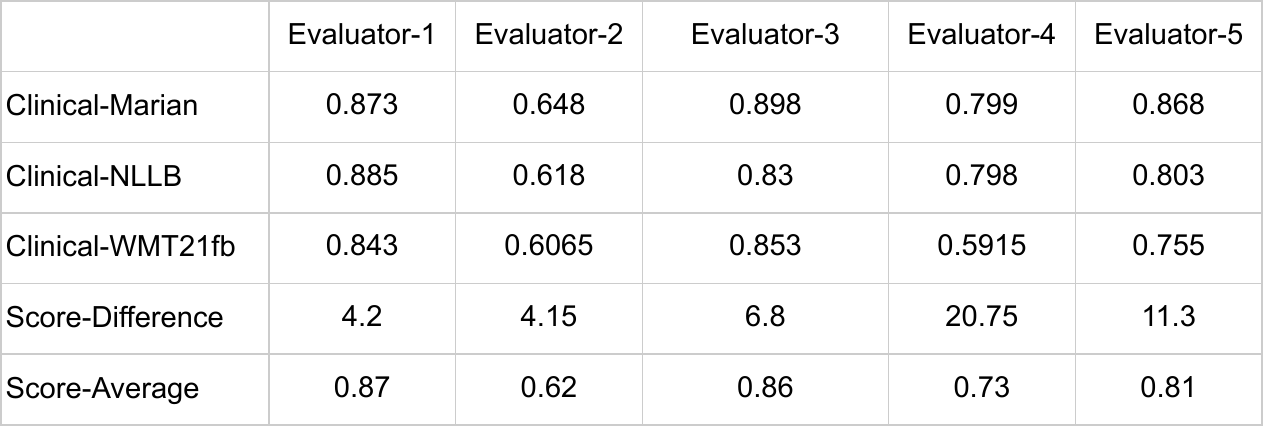}
\caption{Summary of Human Expert-Based Evaluations}
\label{fig:human-eval-sum-cropped}
\end{center}
\end{figure*}

To measure the inter-rater-reliability (IRR) of the human evaluation we carried out,  we summarise the evaluation output from five human evaluators on three models in Figure \ref{fig:human-eval-sum-cropped}. The summaries include the average scores for each model, the score difference between these three models, and the average scores from the three models, from each person.

In this case we have continuous ratings (ranging from 0 to 1) rather than categorical ratings. Therefore, Cohen's Kappa or Fleiss' Kappa are not the most appropriate measures for this work. 
The Intraclass Correlation Coefficient (ICC) which measures  reliability of ratings by comparing  variability of different ratings of the same subject to the total variation across all ratings and all subjects would also not be appropriate here because there is a greater variation within the ratings of the same MT engine than between different MT engines.
 
However, we can compute standard deviations of the evaluations by different reviewers for each engine as follows: 
\begin{itemize}
    \item Marian: approximately 0.101
    \item NLLB: approximately 0.100
    \item WMT21: approximately 0.125
\end{itemize}
 
These values represent the amount of variability in the ratings given by different reviewers for each engine. The confidence intervals for these measurements for confidence level of 80\% are:
\begin{itemize}
    \item Marian: approximately (0.759, 0.875)
    \item NLLB: approximately (0.729, 0.844)
    \item WMT21: approximately (0.658, 0.802)
\end{itemize} 
 
In other words, with 80\% confidence: 
\begin{itemize}
    \item Marian: 0.817 ± 0.058
    \item NLLB: 0.7865 ± 0.0575
    \item WMT21: 0.73 ± 0.072
\end{itemize}

\begin{figure*}[!t]
\begin{center}
\centering
\includegraphics*[width=0.90\textwidth]{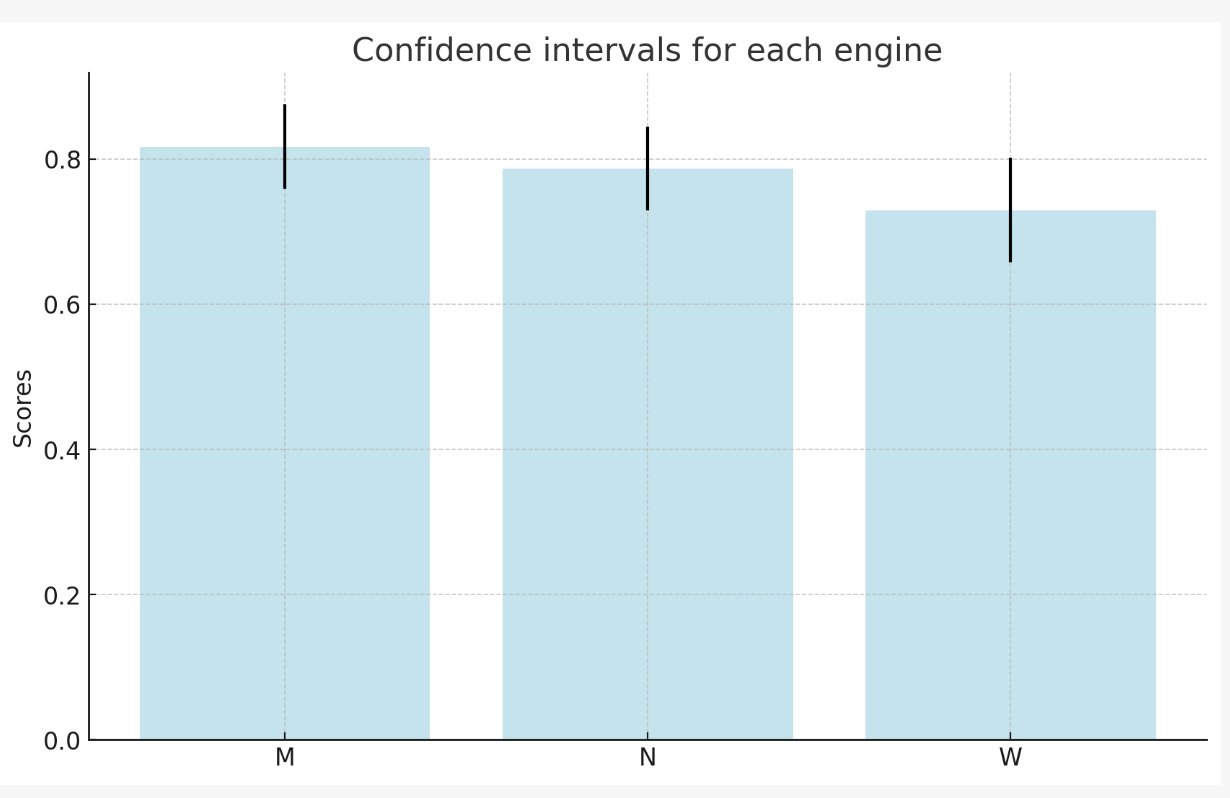}
\caption{Confidence Intervals of Three Models (M, N, W): Clinical-Marian, Clinical-NLLB, and Clinical-WMT21fb }
\label{fig:CI_each_model}
\end{center}
\end{figure*}

This can be visualised in Figure \ref{fig:CI_each_model}. These intervals indeed overlap; however, Marian is reliably better than NLLB, and it is of course extremely surprising that WMT21fb rating is that high, considering that this result has been achieved with transfer learning by fine-tuning the engine without English-Spanish in the original PLM training dataset! As we can see, for some reviewers who are quite tolerant to errors (e.g. Evaluator-1) the quality of all the engines is almost the same.
 The more proficient and knowledgeable the reviewer is, the higher is the difference in their ratings.

\subsection{Error Analysis}

\begin{figure*}[!th]
\centering
\includegraphics*[width=0.99\textwidth]{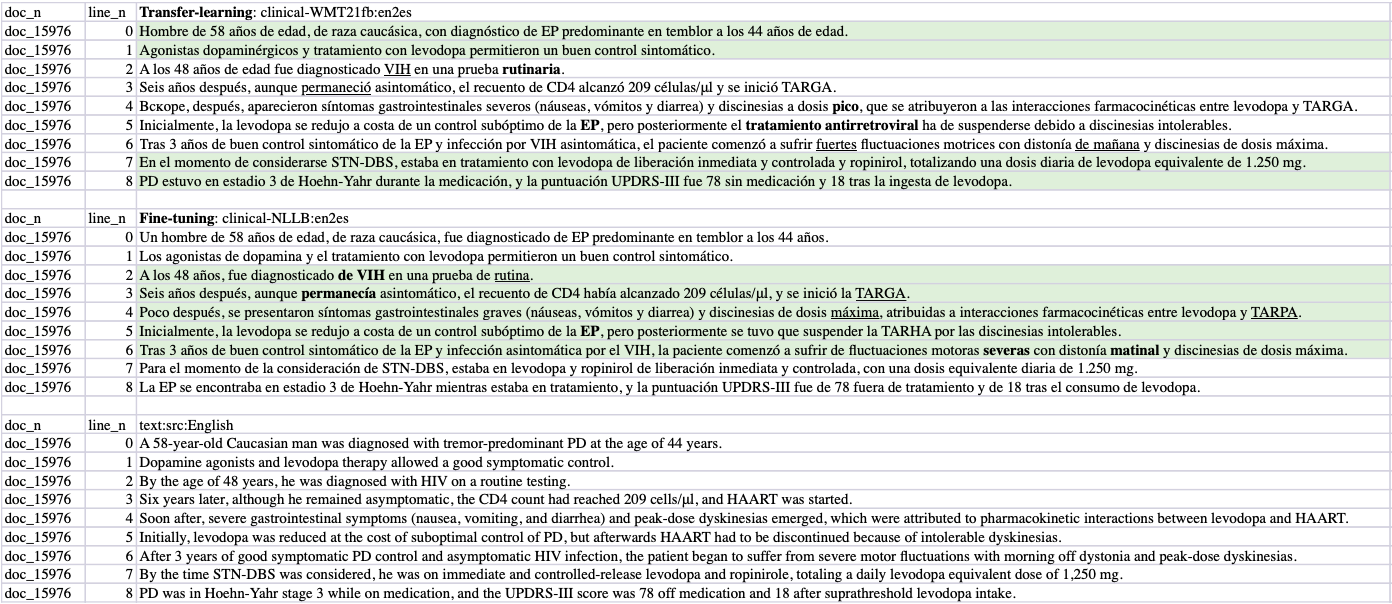}
\caption{ Task-1 Cases/Sentences EN-ES Translation Examples: clinic-WMT21fb \textit{vs} clinic-NLLB}
\label{fig:cases_task_example}
\end{figure*}

\begin{figure*}[!th]
\centering
\includegraphics*[width=0.99\textwidth]{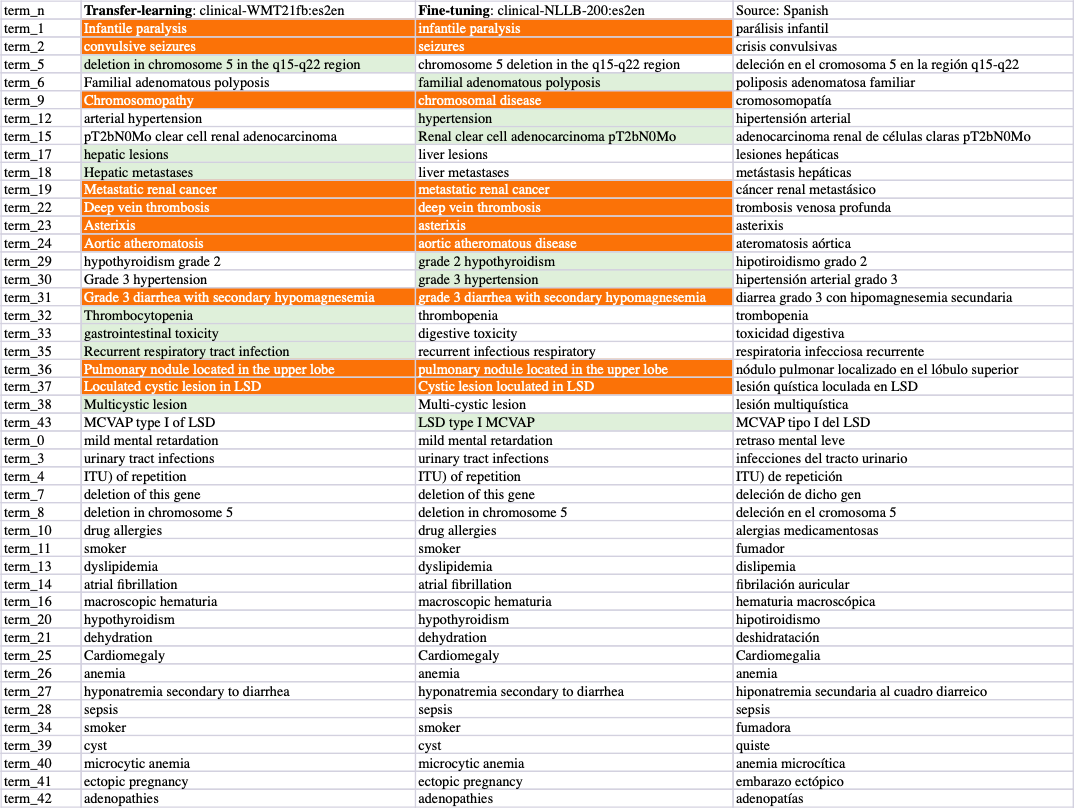}
\caption{ Task-2 Clinical Term ES-EN Translation Examples: clinic-WMT21fb \textit{vs} clinic-NLLB}
\label{fig:term_task_example}
\end{figure*}

\begin{figure*}[!th]
\centering
\includegraphics*[width=0.99\textwidth]{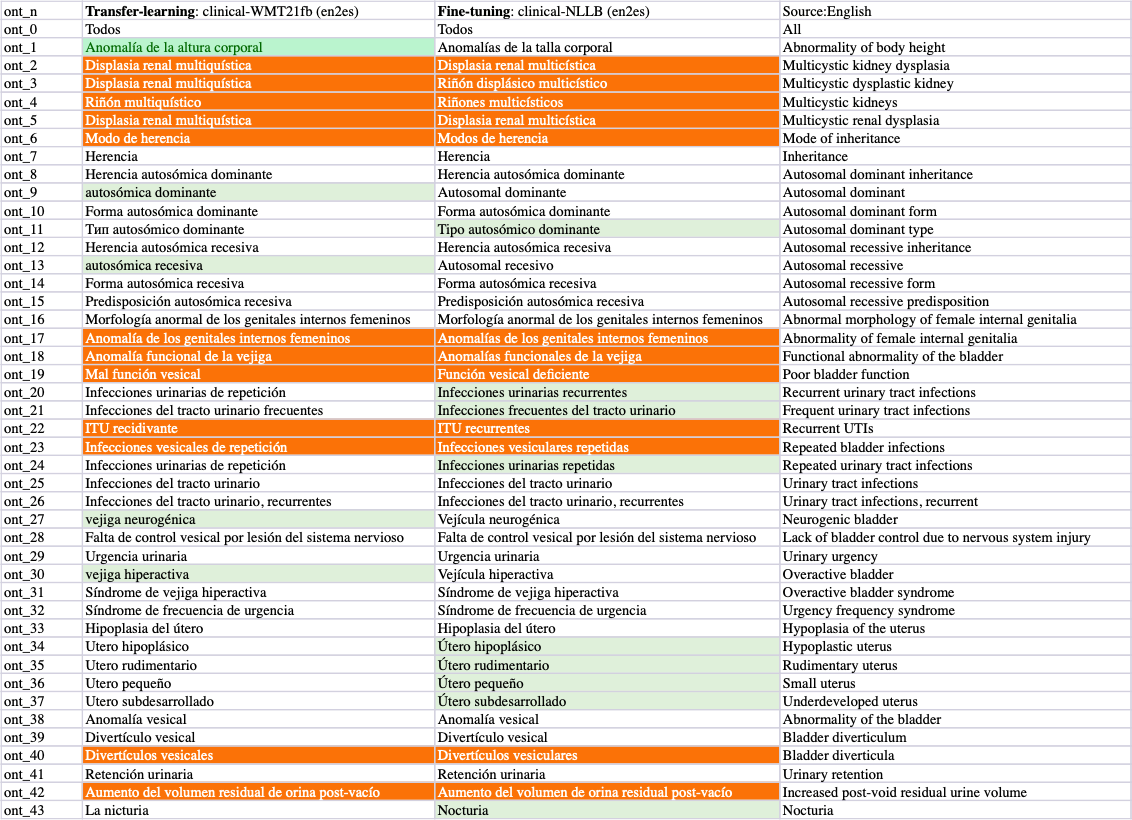}
\caption{ Task-3 Concept EN-ES Translation Examples: clinic-WMT21fb \textit{vs} clinic-NLLB}
\label{fig:concept_task_example}
\end{figure*}

We list sampled error analyses on the outputs from the fine-tuned WMT21fb and NLLB models in Figure \ref{fig:cases_task_example}, \ref{fig:term_task_example}, and \ref{fig:concept_task_example} for the three tasks on translations of sentences, terms, and concepts. The preferred translations are highlighted in green colour and ``both sounds ok'' is marked in orange.

From the comparisons of sampled output sentences, we discovered that the most frequent errors in a fine-grained analysis include \textit{literal} translations, \textit{oral vs written} languages, translation \textit{inconsistency}, \textit{inaccuracy} of terms, \textit{hallucination/made-up} words, and \textit{gender}-related errors such as feminine vs masculine, in addition to the standard fluency and adequacy that have been commonly used by traditional MT researchers \cite{han-etal-2021-translation}. 
For instance, in Figure \ref{fig:cases_task_example}, the first two sentences (line 0 and 1) from clinical-WMT21fb model are more written Spanish than the clinical-NLLB model whose outputs are more oral Spanish.
However, line 6 from clinical-WMT21fb model includes the words ``fuertes'' which means ``strong'' that is not as accurate as ``severas/severe'' from the other model. In addition, ``de manana'' in the same line is less natural than ``matinal'' from clinical-NLLB.
Regarding gender-related issues, we can see the examples also in line 6, where clinical-WMT21fb produced ``el paciente'' in masculine while clinical-NLLB produced ``la paciente'' in feminine. However, the source did not say what gender is ``the patient''. 
Regarding literal translation examples, we can see in Figure \ref{fig:concept_task_example}, line ont-19 shows that clinical-WMT21fb gives more literal translation ``Mal función vesical'' than the preferred one ``Función vesical deficiente'' by clinical-NLLB when translating ``Poor bladder function''.
The neural model output hallucinations can also be found in Figure \ref{fig:concept_task_example}, e.g. ``Vejícula'' does not exist and is likely a mix of ``vejiga'' and ``vesicula'' in Line ont-27; similarly, in Line ont-2, ``multicística'' is a mix of Spanish and English, because the correct Spanish shall be ``multiquística''.

As we mentioned in Section \ref{model_eval_section}, there are 4\% Russian tokens in the English-to-Spanish translation outputs from the Clinical-WMT21fb model which can be observed in Figures \ref{fig:cases_task_example} and \ref{fig:concept_task_example}. However, they are meaningful tokens, not some nonsense, e.g. the Russian tokens in Figure \ref{fig:cases_task_example} from line n-4 means ``soon'' and in Figure \ref{fig:concept_task_example} means ``type of'' from ont-11.

\section{Discussions and Conclusions}
\label{discussion_conclusions}
To boost the knowledge transfer for digital healthcare and get the most knowledge out of available clinical resources, we explored the state-of-the-art neural language models regarding their performances in clinical machine translation.
We investigated a smaller multilingual pre-trained language model (s-MPLM) Marian from the Helsinki NLP group, in comparison to two extra-large MPLM (xL-MPLM) NLLB and WMT21fb from Meta-AI. 
We also investigated the transfer-learning possibility in clinical domain translation using xL-MPLM WMT21fb.
We carried out data cleaning and fine-tuning in the clinical domain. We evaluated our work using both automatic evaluation metrics and human expert-based evaluation using the HOPE \cite{gladkoff-han-2022-hope} framework.


The experiment has led to some far-reaching conclusions about MT models and their design, test, and applications, in particular:

1) \textbf{The bigger size of the model does not mean that the quality is better}. This premise proved to be false, evidently because researchers need vast amounts of data to train very large models and very often such data is not clear enough. 
On the contrary, when we clean the data very well for fine-tuning, we can bring the model quality to much higher levels in specific domains, e.g. clinical text. 
We reached the point where \textit{the data quality was more important than the model's size}.
One key takeaway for researchers and practitioners from this is that if they can get 250,000 clean segments in a new low-resource language, they can  fine-tune large language models (LLMs) and get a good enough engine in this language. Then, the next step is to continue to get clean data by post-editing translation output from that engine. This is a very important implication for ``low resource languages.''

2) \textbf{The automated metrics deliver an illusion of measurement} – they are a good tool for iterative stochastic gradient descent during training, but \textit{they do not measure quality} (only some sort of similarity), are not compatible when any of the underlying factors change, provide results on a non-uniform scale even on their interval of validity,  and in general are not sufficiently reliable, and may be misleading. \textit{We cannot rely on automatic metrics alone}. Instead, human translation quality validation is a must and such validation can deny and reverse the results of automatic measurement.

\section*{Acknowledgements}
The human evaluation of this work was carried out by Betty Galiano, Marta Martínez Albaladejo,	Valeria López Expósito, Carlos Mateos, and 
Alfredo Madrid. We thank our human evaluators for their volunteering and hard work.
We also thank Cristina Sánchez for assisting with double checking the sampled human evaluations.
LH and GN are grateful for the support from the grant “Assembling the Data Jigsaw: Powering Robust Research on the
Causes, Determinants and Outcomes of MSK Disease.” The project has been funded by the Nuffield
Foundation, but the views expressed are those of the authors and not necessarily of the Foundation. 
Visit www.nuffieldfoundation.org. 
LH and GN are also supported by the grant “Integrating hospital outpatient letters into the healthcare data space” (EP/V047949/1; funder: UKRI/EPSRC).


\bibliographystyle{coling}
\bibliography{Frontiers2021}

\begin{thebibliography}{}

\bibitem[\protect\citename{Almansor and Al-Ani}2018]{Almansor2018NMT_lowRes}
Ebtesam~H. Almansor and Ahmed Al-Ani.
\newblock 2018.
\newblock A hybrid neural machine translation technique for translating low resource languages.
\newblock In Petra Perner, editor, {\em Machine Learning and Data Mining in Pattern Recognition}, pages 347--356, Cham. Springer International Publishing.

\bibitem[\protect\citename{Alyafeai \bgroup et al.\egroup }2020]{alyafeai2020survey_TransferL}
Zaid Alyafeai, Maged~Saeed AlShaibani, and Irfan Ahmad.
\newblock 2020.
\newblock A survey on transfer learning in natural language processing.
\newblock {\em arXiv preprint arXiv:2007.04239}.

\bibitem[\protect\citename{Banerjee and Lavie}2005]{BanerjeeLavie2005}
Satanjeev Banerjee and Alon Lavie.
\newblock 2005.
\newblock Meteor: An automatic metric for mt evaluation with improved correlation with human judgments.
\newblock In {\em Proceedings of the ACL}.

\bibitem[\protect\citename{Belkadi \bgroup et al.\egroup }2023]{belkadi2023generating}
Samuel Belkadi, Nicolo Micheletti, Lifeng Han, Warren Del-Pinto, and Goran Nenadic.
\newblock 2023.
\newblock Generating medical instructions with conditional transformer.
\newblock In {\em NeurIPS 2023 Workshop on Synthetic Data Generation with Generative AI}.

\bibitem[\protect\citename{Bodenreider}2004]{bodenreider2004unified_UMLS}
Olivier Bodenreider.
\newblock 2004.
\newblock The unified medical language system (umls): integrating biomedical terminology.
\newblock {\em Nucleic acids research}, 32(suppl\_1):D267--D270.

\bibitem[\protect\citename{Bojar \bgroup et al.\egroup }2016]{bojar-etal-2016-findings_wmt}
Ond{\v{r}}ej Bojar, Rajen Chatterjee, Christian Federmann, Yvette Graham, Barry Haddow, Matthias Huck, Antonio Jimeno~Yepes, Philipp Koehn, Varvara Logacheva, Christof Monz, Matteo Negri, Aur{\'e}lie N{\'e}v{\'e}ol, Mariana Neves, Martin Popel, Matt Post, Raphael Rubino, Carolina Scarton, Lucia Specia, Marco Turchi, Karin Verspoor, and Marcos Zampieri.
\newblock 2016.
\newblock Findings of the 2016 conference on machine translation.
\newblock In {\em Proceedings of the First Conference on Machine Translation: Volume 2, Shared Task Papers}, pages 131--198, Berlin, Germany, August. Association for Computational Linguistics.

\bibitem[\protect\citename{Costa-juss{\`a} \bgroup et al.\egroup }2022]{costa2022no_NLLB}
Marta~R Costa-juss{\`a}, James Cross, Onur {\c{C}}elebi, Maha Elbayad, Kenneth Heafield, Kevin Heffernan, Elahe Kalbassi, Janice Lam, Daniel Licht, Jean Maillard, et~al.
\newblock 2022.
\newblock No language left behind: Scaling human-centered machine translation.
\newblock {\em arXiv preprint arXiv:2207.04672}.

\bibitem[\protect\citename{Cui \bgroup et al.\egroup }2023]{cui-etal-2023-medtem2}
Yang Cui, Lifeng Han, and Goran Nenadic.
\newblock 2023.
\newblock {M}ed{T}em2.0: Prompt-based temporal classification of treatment events from discharge summaries.
\newblock In Vishakh Padmakumar, Gisela Vallejo, and Yao Fu, editors, {\em Proceedings of the 61st Annual Meeting of the Association for Computational Linguistics (Volume 4: Student Research Workshop)}, pages 160--183, Toronto, Canada, July. Association for Computational Linguistics.

\bibitem[\protect\citename{Dehghan \bgroup et al.\egroup }2015]{dehghan2015combining_de-id}
Azad Dehghan, Aleksandar Kovacevic, George Karystianis, John~A Keane, and Goran Nenadic.
\newblock 2015.
\newblock Combining knowledge-and data-driven methods for de-identification of clinical narratives.
\newblock {\em Journal of biomedical informatics}, 58(Suppl):S53.

\bibitem[\protect\citename{Devlin \bgroup et al.\egroup }2018]{bert2018devlin}
Jacob Devlin, Ming{-}Wei Chang, Kenton Lee, and Kristina Toutanova.
\newblock 2018.
\newblock {BERT:} pre-training of deep bidirectional transformers for language understanding.
\newblock {\em CoRR}, abs/1810.04805.

\bibitem[\protect\citename{Dew \bgroup et al.\egroup }2018]{dew2018development_MT_health}
Kristin~N Dew, Anne~M Turner, Yong~K Choi, Alyssa Bosold, and Katrin Kirchhoff.
\newblock 2018.
\newblock Development of machine translation technology for assisting health communication: A systematic review.
\newblock {\em Journal of biomedical informatics}, 85:56--67.

\bibitem[\protect\citename{Donnelly}2006]{donnelly2006snomed}
Kevin Donnelly.
\newblock 2006.
\newblock Snomed-ct: The advanced terminology and coding system for ehealth.
\newblock {\em Studies in health technology and informatics}, 121:279.

\bibitem[\protect\citename{Elbattah \bgroup et al.\egroup }2021]{elbattah2021role_health_review}
Mahmoud Elbattah, {\'E}milien Arnaud, Maxime Gignon, and Gilles Dequen.
\newblock 2021.
\newblock The role of text analytics in healthcare: A review of recent developments and applications.
\newblock {\em Healthinf}, 5:825--832.

\bibitem[\protect\citename{Fan \bgroup et al.\egroup }2021]{10.5555/3546258.3546365_m2m_100}
Angela Fan, Shruti Bhosale, Holger Schwenk, Zhiyi Ma, Ahmed El-Kishky, Siddharth Goyal, Mandeep Baines, Onur Celebi, Guillaume Wenzek, Vishrav Chaudhary, Naman Goyal, Tom Birch, Vitaliy Liptchinsky, Sergey Edunov, Edouard Grave, Michael Auli, and Armand Joulin.
\newblock 2021.
\newblock Beyond english-centric multilingual machine translation.
\newblock {\em J. Mach. Learn. Res.}, 22(1):1--48, jan.

\bibitem[\protect\citename{Finley \bgroup et al.\egroup }2018]{finley-etal-2018-dictations_clinicalMT}
Gregory Finley, Wael Salloum, Najmeh Sadoughi, Erik Edwards, Amanda Robinson, Nico Axtmann, Michael Brenndoerfer, Mark Miller, and David Suendermann-Oeft.
\newblock 2018.
\newblock From dictations to clinical reports using machine translation.
\newblock In {\em Proceedings of the 2018 Conference of the North {A}merican Chapter of the Association for Computational Linguistics: Human Language Technologies, Volume 3 (Industry Papers)}, pages 121--128, New Orleans - Louisiana, June. Association for Computational Linguistics.

\bibitem[\protect\citename{Gladkoff and Han}2022]{gladkoff-han-2022-hope}
Serge Gladkoff and Lifeng Han.
\newblock 2022.
\newblock {HOPE}: A task-oriented and human-centric evaluation framework using professional post-editing towards more effective {MT} evaluation.
\newblock In {\em Proceedings of the Thirteenth Language Resources and Evaluation Conference}, pages 13--21, Marseille, France, June. European Language Resources Association.

\bibitem[\protect\citename{Griciūtė \bgroup et al.\egroup }2023]{griciūtė2023topic}
Bernadeta Griciūtė, Lifeng Han, and Goran Nenadic.
\newblock 2023.
\newblock Topic modelling of swedish newspaper articles about coronavirus: a case study using latent dirichlet allocation method.
\newblock In {\em 2023 IEEE 11th International Conference on Healthcare Informatics (ICHI)}, pages 627--636.

\bibitem[\protect\citename{Han and Kuang}2018]{HanKuang2018NMT}
Lifeng Han and Shaohui Kuang.
\newblock 2018.
\newblock Incorporating chinese radicals into neural machine translation: Deeper than character level.
\newblock In {\em Proceedings of ESSLLI-2018}, pages 54--65. Association for Logic, Language and Information (FoLLI), August.

\bibitem[\protect\citename{Han \bgroup et al.\egroup }2021a]{han-etal-2021-chinese}
Lifeng Han, Gareth Jones, Alan Smeaton, and Paolo Bolzoni.
\newblock 2021a.
\newblock {C}hinese character decomposition for neural {MT} with multi-word expressions.
\newblock In {\em Proceedings of the 23rd Nordic Conference on Computational Linguistics (NoDaLiDa)}, pages 336--344, Reykjavik, Iceland (Online), May 31--2 June. Link{\"o}ping University Electronic Press, Sweden.

\bibitem[\protect\citename{Han \bgroup et al.\egroup }2021b]{han-etal-2021-translation}
Lifeng Han, Alan Smeaton, and Gareth Jones.
\newblock 2021b.
\newblock Translation quality assessment: A brief survey on manual and automatic methods.
\newblock In Yuri Bizzoni, Elke Teich, Cristina Espa{\~n}a-Bonet, and Josef van Genabith, editors, {\em Proceedings for the First Workshop on Modelling Translation: Translatology in the Digital Age}, pages 15--33, online, May. Association for Computational Linguistics.

\bibitem[\protect\citename{Han \bgroup et al.\egroup }2022a]{han-etal-2022-examining}
Lifeng Han, Gleb Erofeev, Irina Sorokina, Serge Gladkoff, and Goran Nenadic.
\newblock 2022a.
\newblock Examining large pre-trained language models for machine translation: What you don{'}t know about it.
\newblock In {\em Proceedings of the Seventh Conference on Machine Translation (WMT)}, pages 908--919, Abu Dhabi, United Arab Emirates (Hybrid), December. Association for Computational Linguistics.

\bibitem[\protect\citename{Han \bgroup et al.\egroup }2022b]{https://doi.org/10.48550/arxiv.2210.06068zero-shotNMTclinical}
Lifeng Han, Gleb Erofeev, Irina Sorokina, Serge Gladkoff, and Goran Nenadic.
\newblock 2022b.
\newblock Using massive multilingual pre-trained language models towards real zero-shot neural machine translation in clinical domain.
\newblock {\em CoRR}, abs/2210.06068.

\bibitem[\protect\citename{Han \bgroup et al.\egroup }2023]{han-etal-2023-investigating}
Lifeng Han, Gleb Erofeev, Irina Sorokina, Serge Gladkoff, and Goran Nenadic.
\newblock 2023.
\newblock Investigating massive multilingual pre-trained machine translation models for clinical domain via transfer learning.
\newblock In Tristan Naumann, Asma Ben~Abacha, Steven Bethard, Kirk Roberts, and Anna Rumshisky, editors, {\em Proceedings of the 5th Clinical Natural Language Processing Workshop}, pages 31--40, Toronto, Canada, July. Association for Computational Linguistics.

\bibitem[\protect\citename{Han}2022]{han2022investigation}
Lifeng Han.
\newblock 2022.
\newblock {\em An investigation into multi-word expressions in machine translation}.
\newblock {Ph.D.} thesis, Dublin City University.

\bibitem[\protect\citename{Henry \bgroup et al.\egroup }2020]{henry20202018_n2c2_task2}
Sam Henry, Kevin Buchan, Michele Filannino, Amber Stubbs, and Ozlem Uzuner.
\newblock 2020.
\newblock 2018 n2c2 shared task on adverse drug events and medication extraction in electronic health records.
\newblock {\em Journal of the American Medical Informatics Association}, 27(1):3--12.

\bibitem[\protect\citename{Islam \bgroup et al.\egroup }2021]{islam2021towards_MT_lowRes}
Md~Adnanul Islam, Md~Saidul~Hoque Anik, and ABM Alim~Al Islam.
\newblock 2021.
\newblock Towards achieving a delicate blending between rule-based translator and neural machine translator.
\newblock {\em Neural Computing and Applications}, 33:12141--12167.

\bibitem[\protect\citename{Jiang \bgroup et al.\egroup }2022]{jiang2022transfer_L_LowResMT}
Hao Jiang, Chao Zhang, Zhihui Xin, Xiaoqiao Huang, Chengli Li, and Yonghang Tai.
\newblock 2022.
\newblock Transfer learning based on lexical constraint mechanism in low-resource machine translation.
\newblock {\em Computers and Electrical Engineering}, 100:107856.

\bibitem[\protect\citename{Junczys-Dowmunt and Grundkiewicz}2017]{junczys-dowmunt-grundkiewicz-2017-exploration_auto-PED}
Marcin Junczys-Dowmunt and Roman Grundkiewicz.
\newblock 2017.
\newblock An exploration of neural sequence-to-sequence architectures for automatic post-editing.
\newblock In {\em Proceedings of the Eighth International Joint Conference on Natural Language Processing (Volume 1: Long Papers)}, pages 120--129, Taipei, Taiwan, November. Asian Federation of Natural Language Processing.

\bibitem[\protect\citename{Junczys-Dowmunt \bgroup et al.\egroup }2018a]{mariannmt}
Marcin Junczys-Dowmunt, Roman Grundkiewicz, Tomasz Dwojak, Hieu Hoang, Kenneth Heafield, Tom Neckermann, Frank Seide, Ulrich Germann, Alham Fikri~Aji, Nikolay Bogoychev, Andr\'{e} F.~T. Martins, and Alexandra Birch.
\newblock 2018a.
\newblock Marian: Fast neural machine translation in {C++}.
\newblock In {\em Proceedings of ACL 2018, System Demonstrations}, pages 116--121, Melbourne, Australia, July. Association for Computational Linguistics.

\bibitem[\protect\citename{Junczys-Dowmunt \bgroup et al.\egroup }2018b]{junczys-dowmunt-etal-2018-marian-cost_WNMTG}
Marcin Junczys-Dowmunt, Kenneth Heafield, Hieu Hoang, Roman Grundkiewicz, and Anthony Aue.
\newblock 2018b.
\newblock {M}arian: Cost-effective high-quality neural machine translation in {C}++.
\newblock In {\em Proceedings of the 2nd Workshop on Neural Machine Translation and Generation}, pages 129--135, Melbourne, Australia, July. Association for Computational Linguistics.

\bibitem[\protect\citename{Khoong and Rodriguez}2022]{khoong2022research_agenda_MT4clinical}
Elaine~C Khoong and Jorge~A Rodriguez.
\newblock 2022.
\newblock A research agenda for using machine translation in clinical medicine.
\newblock {\em Journal of General Internal Medicine}, 37(5):1275--1277.

\bibitem[\protect\citename{Kova{\v{c}}evi{\'c} \bgroup et al.\egroup }2013]{kovavcevic2013combining_clinical}
Aleksandar Kova{\v{c}}evi{\'c}, Azad Dehghan, Michele Filannino, John~A Keane, and Goran Nenadic.
\newblock 2013.
\newblock Combining rules and machine learning for extraction of temporal expressions and events from clinical narratives.
\newblock {\em Journal of the American Medical Informatics Association: JAMIA}, 20(5):859.

\bibitem[\protect\citename{Kuang \bgroup et al.\egroup }2018]{kuang-etal-2018-attentionNMT}
Shaohui Kuang, Junhui Li, Ant{\'o}nio Branco, Weihua Luo, and Deyi Xiong.
\newblock 2018.
\newblock Attention focusing for neural machine translation by bridging source and target embeddings.
\newblock In {\em Proceedings of the 56th Annual Meeting of the Association for Computational Linguistics (Volume 1: Long Papers)}, pages 1767--1776, Melbourne, Australia, July. Association for Computational Linguistics.

\bibitem[\protect\citename{Lepikhin \bgroup et al.\egroup }2020]{lepikhingshard_Scaling}
Dmitry Lepikhin, HyoukJoong Lee, Yuanzhong Xu, Dehao Chen, Orhan Firat, Yanping Huang, Maxim Krikun, Noam Shazeer, and Zhifeng Chen.
\newblock 2020.
\newblock Gshard: Scaling giant models with conditional computation and automatic sharding.
\newblock In {\em International Conference on Learning Representations}.

\bibitem[\protect\citename{Lin}2004]{lin-2004-rouge}
Chin-Yew Lin.
\newblock 2004.
\newblock {ROUGE}: A package for automatic evaluation of summaries.
\newblock In {\em Text Summarization Branches Out}, pages 74--81, Barcelona, Spain, July. Association for Computational Linguistics.

\bibitem[\protect\citename{Luo \bgroup et al.\egroup }2022]{DL4clinical_social}
Xiao Luo, Priyanka Gandhi, Susan Storey, and Kun Huang.
\newblock 2022.
\newblock A deep language model for symptom extraction from clinical text and its application to extract covid-19 symptoms from social media.
\newblock {\em IEEE Journal of Biomedical and Health Informatics}, 26(4):1737--1748.

\bibitem[\protect\citename{Manchanda and Bhagwat}2022]{manchanda-bhagwat-2022-optums}
Sahil Manchanda and Saurabh Bhagwat.
\newblock 2022.
\newblock Optum{'}s submission to {WMT}22 biomedical translation tasks.
\newblock In {\em Proceedings of the Seventh Conference on Machine Translation (WMT)}, pages 925--929, Abu Dhabi, United Arab Emirates (Hybrid), December. Association for Computational Linguistics.

\bibitem[\protect\citename{Mujjiga \bgroup et al.\egroup }2019]{mujjiga2019identifying_semantic_MT_clinical}
Srikanth Mujjiga, Vamsi Krishna, Kalyan Chakravarthi, and J~Vijayananda.
\newblock 2019.
\newblock Identifying semantics in clinical reports using neural machine translation.
\newblock In {\em Proceedings of the AAAI conference on artificial intelligence}, volume~33, pages 9552--9557.

\bibitem[\protect\citename{Neves \bgroup et al.\egroup }2022]{neves-etal-2022-findings_biomedMT}
Mariana Neves, Antonio Jimeno~Yepes, Amy Siu, Roland Roller, Philippe Thomas, Maika Vicente~Navarro, Lana Yeganova, Dina Wiemann, Giorgio~Maria Di~Nunzio, Federica Vezzani, Christel Gerardin, Rachel Bawden, Darryl~Johan Estrada, Salvador Lima-lopez, Eulalia Farre-maduel, Martin Krallinger, Cristian Grozea, and Aurelie Neveol.
\newblock 2022.
\newblock Findings of the {WMT} 2022 biomedical translation shared task: Monolingual clinical case reports.
\newblock In {\em Proceedings of the Seventh Conference on Machine Translation (WMT)}, pages 694--723, Abu Dhabi, United Arab Emirates (Hybrid), December. Association for Computational Linguistics.

\bibitem[\protect\citename{Nguyen \bgroup et al.\egroup }2023]{nguyen2023spanbased_NER}
Nhung T.~H. Nguyen, Makoto Miwa, and Sophia Ananiadou.
\newblock 2023.
\newblock Span-based named entity recognition by generating and compressing information.
\newblock In Andreas Vlachos and Isabelle Augenstein, editors, {\em Proceedings of the 17th Conference of the European Chapter of the Association for Computational Linguistics}, pages 1984--1996, Dubrovnik, Croatia, May. Association for Computational Linguistics.

\bibitem[\protect\citename{Noor \bgroup et al.\egroup }2022]{noor2022deployment_TA_hospital}
Kawsar Noor, Lukasz Roguski, Xi~Bai, Alex Handy, Roman Klapaukh, Amos Folarin, Luis Romao, Joshua Matteson, Nathan Lea, Leilei Zhu, et~al.
\newblock 2022.
\newblock Deployment of a free-text analytics platform at a uk national health service research hospital: Cogstack at university college london hospitals.
\newblock {\em JMIR Medical Informatics}, 10(8):e38122.

\bibitem[\protect\citename{Oyebode \bgroup et al.\egroup }2021]{oyebode2021health_social}
Oladapo Oyebode, Chinenye Ndulue, Ashfaq Adib, Dinesh Mulchandani, Banuchitra Suruliraj, Fidelia~Anulika Orji, Christine~T Chambers, Sandra Meier, Rita Orji, et~al.
\newblock 2021.
\newblock Health, psychosocial, and social issues emanating from the covid-19 pandemic based on social media comments: text mining and thematic analysis approach.
\newblock {\em JMIR medical informatics}, 9(4):e22734.

\bibitem[\protect\citename{Papineni \bgroup et al.\egroup }2002]{papineni-etal-2002-bleu}
Kishore Papineni, Salim Roukos, Todd Ward, and Wei-Jing Zhu.
\newblock 2002.
\newblock {B}leu: a method for automatic evaluation of machine translation.
\newblock In {\em Proceedings of the 40th Annual Meeting of the Association for Computational Linguistics}, pages 311--318, Philadelphia, Pennsylvania, USA, July. Association for Computational Linguistics.

\bibitem[\protect\citename{Peng \bgroup et al.\egroup }2019]{peng-etal-2019-transfer_biomedNLP}
Yifan Peng, Shankai Yan, and Zhiyong Lu.
\newblock 2019.
\newblock Transfer learning in biomedical natural language processing: An evaluation of {BERT} and {ELM}o on ten benchmarking datasets.
\newblock In Dina Demner-Fushman, Kevin~Bretonnel Cohen, Sophia Ananiadou, and Junichi Tsujii, editors, {\em Proceedings of the 18th BioNLP Workshop and Shared Task}, pages 58--65, Florence, Italy, August. Association for Computational Linguistics.

\bibitem[\protect\citename{Percha}2021]{percha2021modern_clinicalTM}
Bethany Percha.
\newblock 2021.
\newblock Modern clinical text mining: a guide and review.
\newblock {\em Annual review of biomedical data science}, 4:165--187.

\bibitem[\protect\citename{Pomares-Quimbaya \bgroup et al.\egroup }2021]{pomares2021transfer_sp_en_classify_clinical}
Alexandra Pomares-Quimbaya, Pilar L{\'o}pez-{\'U}beda, and Stefan Schulz.
\newblock 2021.
\newblock Transfer learning for classifying spanish and english text by clinical specialties.
\newblock In {\em Public Health and Informatics}, pages 377--381. IOS Press.

\bibitem[\protect\citename{Post}2018]{post-2018-call4clarity}
Matt Post.
\newblock 2018.
\newblock A call for clarity in reporting {BLEU} scores.
\newblock In {\em Proceedings of the Third Conference on Machine Translation: Research Papers}, pages 186--191, Belgium, Brussels, October. Association for Computational Linguistics.

\bibitem[\protect\citename{Qian \bgroup et al.\egroup }2021]{qian2021cpas_UK_ML_hospital}
Zhaozhi Qian, Ahmed~M Alaa, and Mihaela van~der Schaar.
\newblock 2021.
\newblock Cpas: the uk’s national machine learning-based hospital capacity planning system for covid-19.
\newblock {\em Machine Learning}, 110:15--35.

\bibitem[\protect\citename{Randhawa \bgroup et al.\egroup }2013]{randhawa2013using_MT_clinical}
Gurdeeshpal Randhawa, Mariella Ferreyra, Rukhsana Ahmed, Omar Ezzat, and Kevin Pottie.
\newblock 2013.
\newblock Using machine translation in clinical practice.
\newblock {\em Canadian Family Physician}, 59(4):382--383.

\bibitem[\protect\citename{Rei \bgroup et al.\egroup }2020]{rei-etal-2020-comet}
Ricardo Rei, Craig Stewart, Ana~C Farinha, and Alon Lavie.
\newblock 2020.
\newblock {COMET}: A neural framework for {MT} evaluation.
\newblock In {\em Proceedings of the 2020 Conference on Empirical Methods in Natural Language Processing (EMNLP)}, pages 2685--2702, Online, November. Association for Computational Linguistics.

\bibitem[\protect\citename{Sennrich \bgroup et al.\egroup }2017]{nematus}
Rico Sennrich, Orhan Firat, Kyunghyun Cho, Alexandra Birch, Barry Haddow, Julian Hitschler, Marcin Junczys-Dowmunt, Samuel L{\"a}ubli, Antonio~Valerio {Miceli Barone}, Jozef Mokry, and Maria Nadejde.
\newblock 2017.
\newblock Nematus: a toolkit for neural machine translation.
\newblock In {\em {Proceedings of the Demonstrations at the 15th Conference of the European Chapter of the Association for Computational Linguistics}}, Valencia, Spain.

\bibitem[\protect\citename{Soto \bgroup et al.\egroup }2019]{soto2019leveraging_SNOMED_ML}
Xabier Soto, Olatz Perez-De-Vinaspre, Maite Oronoz, and Gorka Labaka.
\newblock 2019.
\newblock Leveraging snomed ct terms and relations for machine translation of clinical texts from basque to spanish.
\newblock In {\em Proceedings of the Second Workshop on Multilingualism at the Intersection of Knowledge Bases and Machine Translation}, pages 8--18.

\bibitem[\protect\citename{Spasic and Nenadic}2020]{spasic2020clinical_review}
Irena Spasic and Goran Nenadic.
\newblock 2020.
\newblock Clinical text data in machine learning: Systematic review.
\newblock {\em JMIR Medical Informatics}, 8(3):e17984.

\bibitem[\protect\citename{Tiedemann and Thottingal}2020]{TiedemannThottingal:EAMT2020_OPUS_MT}
J{\"o}rg Tiedemann and Santhosh Thottingal.
\newblock 2020.
\newblock {OPUS-MT} — {B}uilding open translation services for the {W}orld.
\newblock In {\em Proceedings of the 22nd Annual Conferenec of the European Association for Machine Translation (EAMT)}, Lisbon, Portugal.

\bibitem[\protect\citename{Tiedemann}2012]{tiedemann2012parallel_OPUS}
J{\"o}rg Tiedemann.
\newblock 2012.
\newblock Parallel data, tools and interfaces in opus.
\newblock In {\em Lrec}, volume 2012, pages 2214--2218. Citeseer.

\bibitem[\protect\citename{Tran \bgroup et al.\egroup }2021]{tran2021facebook}
Chau Tran, Shruti Bhosale, James Cross, Philipp Koehn, Sergey Edunov, and Angela Fan.
\newblock 2021.
\newblock Facebook ai’s wmt21 news translation task submission.
\newblock In {\em Proc. of WMT}.

\bibitem[\protect\citename{Tu \bgroup et al.\egroup }2023]{tu2023extraction}
Hangyu Tu, Lifeng Han, and Goran Nenadic.
\newblock 2023.
\newblock Extraction of medication and temporal relation from clinical text using neural language models.
\newblock In {\em 2023 IEEE International Conference on Big Data (BigData)}, pages 2735--2744. IEEE.

\bibitem[\protect\citename{Vaswani \bgroup et al.\egroup }2017]{google2017attention}
Ashish Vaswani, Noam Shazeer, Niki Parmar, Jakob Uszkoreit, Llion Jones, Aidan~N. Gomez, Lukasz Kaiser, and Illia Polosukhin.
\newblock 2017.
\newblock Attention is all you need.
\newblock In {\em Proceedings of the 31st Conference on Neural Information Processing System}, pages 6000--6010, Long Beach, CA, USA.

\bibitem[\protect\citename{Villegas \bgroup et al.\egroup }2018]{villegas2018mespen}
Marta Villegas, Ander Intxaurrondo, Aitor Gonzalez-Agirre, Montserrat Marimon, and Martin Krallinger.
\newblock 2018.
\newblock The mespen resource for english-spanish medical machine translation and terminologies: census of parallel corpora, glossaries and term translations.
\newblock {\em LREC MultilingualBIO: multilingual biomedical text processing}.

\bibitem[\protect\citename{Wang \bgroup et al.\egroup }2022]{wang-etal-2022-huawei}
Weixuan Wang, Xupeng Meng, Suqing Yan, Ye~Tian, and Wei Peng.
\newblock 2022.
\newblock Huawei {B}abel{T}ar {NMT} at {WMT}22 biomedical translation task: How we further improve domain-specific {NMT}.
\newblock In {\em Proceedings of the Seventh Conference on Machine Translation (WMT)}, pages 930--935, Abu Dhabi, United Arab Emirates (Hybrid), December. Association for Computational Linguistics.

\bibitem[\protect\citename{Weaver}1955]{Weaver1955}
Warren Weaver.
\newblock 1955.
\newblock Translation.
\newblock {\em Machine Translation of Languages: Fourteen Essays}.

\bibitem[\protect\citename{Wroge \bgroup et al.\egroup }2018]{wroge2018parkinson_diagnosis_ML}
Timothy~J Wroge, Yasin {\"O}zkanca, Cenk Demiroglu, Dong Si, David~C Atkins, and Reza~Hosseini Ghomi.
\newblock 2018.
\newblock Parkinson’s disease diagnosis using machine learning and voice.
\newblock In {\em 2018 IEEE signal processing in medicine and biology symposium (SPMB)}, pages 1--7. IEEE.

\bibitem[\protect\citename{Wu \bgroup et al.\egroup }2022]{wu2022cross_PLM_clinical}
Yuping Wu, Lifeng Han, Valerio Antonini, and Goran Nenadic.
\newblock 2022.
\newblock On cross-domain pre-trained language models for clinical text mining: How do they perform on data-constrained fine-tuning?
\newblock {\em arXiv preprint arXiv:2210.12770}.

\bibitem[\protect\citename{Yang \bgroup et al.\egroup }2009]{yang2009text_clinical}
Hui Yang, Irena Spasic, John~A Keane, and Goran Nenadic.
\newblock 2009.
\newblock A text mining approach to the prediction of disease status from clinical discharge summaries.
\newblock {\em Journal of the American Medical Informatics Association: JAMIA}, 16(4):596.

\bibitem[\protect\citename{Yeganova \bgroup et al.\egroup }2021]{yeganova-etal-2021-findings_biomedMT}
Lana Yeganova, Dina Wiemann, Mariana Neves, Federica Vezzani, Amy Siu, Inigo Jauregi~Unanue, Maite Oronoz, Nancy Mah, Aur{\'e}lie N{\'e}v{\'e}ol, David Martinez, Rachel Bawden, Giorgio~Maria Di~Nunzio, Roland Roller, Philippe Thomas, Cristian Grozea, Olatz Perez-de Vi{\~n}aspre, Maika Vicente~Navarro, and Antonio Jimeno~Yepes.
\newblock 2021.
\newblock Findings of the {WMT} 2021 biomedical translation shared task: Summaries of animal experiments as new test set.
\newblock In {\em Proceedings of the Sixth Conference on Machine Translation}, pages 664--683, Online, November. Association for Computational Linguistics.

\bibitem[\protect\citename{Zhang \bgroup et al.\egroup }2021]{zhang2021share_schedule}
Biao Zhang, Ankur Bapna, Rico Sennrich, and Orhan Firat.
\newblock 2021.
\newblock Share or not? learning to schedule language-specific capacity for multilingual translation.
\newblock In {\em Ninth International Conference on Learning Representations 2021}.

\bibitem[\protect\citename{Zhu \bgroup et al.\egroup }2021]{zhu2021classification_covid_DL}
Ziwei Zhu, Zhang Xingming, Guihua Tao, Tingting Dan, Jiao Li, Xijie Chen, Yang Li, Zhichao Zhou, Xiang Zhang, Jinzhao Zhou, et~al.
\newblock 2021.
\newblock Classification of covid-19 by compressed chest ct image through deep learning on a large patients cohort.
\newblock {\em Interdisciplinary Sciences: Computational Life Sciences}, 13:73--82.

\end{thebibliography}

\section*{Appendix}


\end{document}